%% file: arxiv.tex
\documentclass[sigconf,nonacm,pdfa]{acmart}

\usepackage[utf8]{inputenc}
\usepackage{amsmath,amsfonts}
\usepackage{amsthm}
\usepackage{textcomp}
\usepackage{color,xcolor}
\usepackage{multirow}
\usepackage{array}
\usepackage{enumitem}
\usepackage{epstopdf}
\usepackage{booktabs}
\usepackage{threeparttable}
\usepackage{bm}
\usepackage{graphicx}
\usepackage{subfigure}
\usepackage{caption}
\usepackage{balance}
\usepackage[a-2b,mathxmp]{pdfx}[2018/12/22] % use the pdfx package to create a PDF/A file from the LATEX source

\newtheoremstyle{mydefn}
{}{}
{\it}       % body font
{0pt}       % indent
{\bfseries} % head font
{:~}        % punctuation after head
{0.25em}    % spacing after head
{}          % CUSTOM-HEAD-SPEC
\theoremstyle{mydefn}

\newtheorem{definition}{Definition}[section]

\newtheoremstyle{myexample}
{}{}
{}          % body font
{0pt}       % indent
{\bfseries} % head font
{:~}        % punctuation after head
{0.25em}    % spacing after head
{}          % CUSTOM-HEAD-SPEC
\theoremstyle{myexample}

\newtheorem{example}{Example}[section]

\renewcommand{\paragraph}[1]{\vspace{0.5em}\noindent\textbf{#1.}}
\renewcommand{\subparagraph}[1]{\vspace{0.5em}\noindent\textbf{\underline{#1.}}}
\renewcommand{\sec}[1]{\textbf{\S#1}}
\newcommand{\keypoint}[1]{\textbf{\underline{#1}}}

\newcommand{\abso}[1]{\left\vert #1 \right\vert}

\newcommand\vldbdoi{XX.XX/XXX.XX}
\newcommand\vldbpages{XXX - XXX}
\newcommand\vldbvolume{17}
\newcommand\vldbissue{3}
\newcommand\vldbyear{2023}
\newcommand\vldbauthors{\authors}
\newcommand\vldbtitle{\shorttitle} 
\newcommand\vldbavailabilityurl{https://github.com/YihaoAng/TSGBench/}
\newcommand\vldbpagestyle{empty}

\begin{document}

\input{01_title}
\input{02_abstract}
\pagestyle{\vldbpagestyle}
\begingroup\small\noindent\raggedright\textbf{PVLDB Reference Format:}\\
\vldbauthors. \vldbtitle. PVLDB, \vldbvolume(\vldbissue): \vldbpages, \vldbyear.\\
\href{https://doi.org/\vldbdoi}{doi:\vldbdoi}
\endgroup
\begingroup
\renewcommand\thefootnote{}\footnote{\noindent
This work is licensed under the Creative Commons BY-NC-ND 4.0 International License. Visit \url{https://creativecommons.org/licenses/by-nc-nd/4.0/} to view a copy of this license. For any use beyond those covered by this license, obtain permission by emailing \href{mailto:info@vldb.org}{info@vldb.org}. Copyright is held by the owner/author(s). Publication rights licensed to the VLDB Endowment. \\
\raggedright Proceedings of the VLDB Endowment, Vol. \vldbvolume, No. \vldbissue\ %
ISSN 2150-8097. \\
\href{https://doi.org/\vldbdoi}{doi:\vldbdoi} \\
}\addtocounter{footnote}{-1}\endgroup
%% VLDB block end %%

%% --------------------------------------------------------------
%% do not modify the following VLDB block
%% --------------------------------------------------------------
%% VLDB block start %%
\ifdefempty{\vldbavailabilityurl}{}{
\vspace{.3cm}
\begingroup\small\noindent\raggedright\textbf{PVLDB Artifact Availability:}\\
The source code, data, and other artifacts have been made available at \url{\vldbavailabilityurl}.
\endgroup
}
%%% VLDB block end %%%

\input{03_introduction}
\input{04_preliminaries}

\input{05_taxonomy}
\input{06_benchmark}
\input{07_setup}
\input{08_experiments}
\input{09_conclusions}

\balance
\bibliographystyle{ACM-Reference-Format}
\bibliography{arxiv}

\end{document}

%% file: 01_title.tex
\title{TSGBench: Time Series Generation Benchmark}

\author{Yihao Ang}
\orcid{0009-0009-1564-4937}
\affiliation{
 \institution{National University of Singapore}
 \institution{NUS Research Institute in Chongqing}
 \country{}
}
\email{yihao\_ang@comp.nus.edu.sg}

\author{Qiang Huang}
\authornote{Qiang Huang is the corresponding author.}
\orcid{0000-0003-1120-4685}
\affiliation{
 \institution{National University of Singapore}
 \country{}
}
\email{huangq@comp.nus.edu.sg}

\author{Yifan Bao}
\orcid{0009-0000-9672-0747}
\affiliation{
 \institution{National University of Singapore}
 \country{}
}
\email{yifan\_bao@comp.nus.edu.sg}

\author{Anthony K. H. Tung}
\orcid{0000-0001-7300-6196}
\affiliation{
 \institution{National University of Singapore}
 \country{}
}
\email{atung@comp.nus.edu.sg}

\author{Zhiyong Huang}
\orcid{0000-0002-1931-7775}
\affiliation{
 \institution{National University of Singapore}
 \institution{NUS Research Institute in Chongqing}
 \country{}
}
\email{huangzy@comp.nus.edu.sg}

%% file: 02_abstract.tex
\begin{abstract}
Synthetic Time Series Generation (TSG) is crucial in a range of applications, including data augmentation, anomaly detection, and privacy preservation. 
Although significant strides have been made in this field, existing methods exhibit three key limitations: 
(1) They often benchmark against similar model types, constraining a holistic view of performance capabilities. 
(2) The use of specialized synthetic and private datasets introduces biases and hampers generalizability. 
(3) Ambiguous evaluation measures, often tied to custom networks or downstream tasks, hinder consistent and fair comparison.

To overcome these limitations, we introduce \textsf{TSGBench}, the inaugural Time Series Generation Benchmark, designed for a unified and comprehensive assessment of TSG methods. It comprises three modules: (1) a curated collection of publicly available, real-world datasets tailored for TSG, together with a standardized preprocessing pipeline; (2) a comprehensive evaluation measures suite including vanilla measures, new distance-based assessments, and visualization tools; (3) a pioneering generalization test rooted in Domain Adaptation (DA), compatible with all methods. 
We have conducted comprehensive experiments using \textsf{TSGBench} across a spectrum of ten real-world datasets from diverse domains, utilizing ten advanced TSG methods and twelve evaluation measures. 
The results highlight the reliability and efficacy of \textsf{TSGBench} in evaluating TSG methods.
Crucially, \textsf{TSGBench} delivers a statistical analysis of the performance rankings of these methods, illuminating their varying performance across different datasets and measures and offering nuanced insights into the effectiveness of each method.
\end{abstract}

\maketitle

%% file: 03_introduction.tex
\section{Introduction}
\label{sec:intro}

%%% Time Series Generation (TSG) task
Within the myriad of tasks centered on time series, synthetic Time Series Generation (TSG) stands out as a burgeoning area of focus due to growing demands in data augmentation \cite{t-cgan}, anomaly detection \cite{cad, campos2021unsupervised}, privacy protection \cite{pategan}, and domain transfer \cite{sasa}.
%%% high-level goal of TSG
TSG aims to produce time series akin to the original, preserving temporal dependencies and dimensional correlations while ensuring the generated time series remains useful for various downstream tasks like classification \cite{li2022ips, ding2022towards} and forecasting \cite{wu2021autocts, cirstea2022towards}.

% TSG methods
Towards this end, numerous methods have emerged to generate synthetic time series. 
Their overarching objective is to develop a generative model that accurately captures the features and dependencies inherent in the input time series, thereby generating new time series that maintain both utility and statistical characteristics.
%%% example methods
For instance, many representative methods \cite{rcgan, timegan, rtsgan, cosci-gan, aec-gan} utilize the power of Generative Adversarial Networks (GANs), integrating distinctive time series architectures to adeptly capture temporal and dimensional dependencies.
Others harness Variational AutoEncoders (VAEs) \cite{timevae, crvae, timevqvae} to strike a balance between data fidelity and the statistical consistency of latent space, thereby enhancing interpretability.
Additionally, some recent approaches employ flow-based models \cite{fourier, gtgan} to provide explicit likelihood modeling, facilitating effective optimization.

\subsection{Motivations}
\label{sec:intro:motivation}

Despite the strides made by these pioneering studies, TSG lags relative to other time series tasks, with three primary limitations (\textbf{L1--L3}) emerging prominently:

% Ad-hoc
\paragraph{L1: A comprehensive taxonomy and comparative analysis of various methods are lacking, limiting a holistic view of performance capabilities}
When selecting baselines for comparison, researchers often choose models from their methodological realm, missing a broad performance comparison across various methodologies.
For example, GT-GAN \cite{gtgan} and AEC-GAN \cite{aec-gan} predominantly consider advanced GAN-based methods, overlooking potential contenders like VAE-based approaches. 
Furthermore, some methods \cite{rtsgan, psa-gan, ls4} mainly target specific downstream tasks such as missing value imputation or forecasting, hindering the application of the generated time series to other tasks.

% Dataset
\paragraph{L2: Inconsistent dataset selection and preprocessing introduce biases and curb generalizability}
%%%
First, the dataset choice greatly influences generation outcomes. 
Different from other time series tasks, the datasets for TSG are diverse due to their lack of label constraints, broadening data source possibilities.
Yet, many studies resort to private and/or synthetic datasets for validation \cite{timegan, timevae, cosci-gan}. 
Private datasets, while varied, hinder reproducibility due to their inaccessibility. 
Moreover, synthetic datasets, albeit straightforward, may oversimplify real-world complexities, risking biased performance evaluations \cite{timegan, cot-gan}.
%%%
Second, preprocessing choices significantly impact generation results.
For instance, many methods \cite{timegan, imitation, gtgan} use a fixed 24-unit sliding window to segment time series, which often inadequately represent their full periods and temporal patterns.
Practices like normalizing time series to a range of $[0,1]$, either before or after applying the sliding window, can lead to discrepancies in results \cite{timegan}. 
In addition, datasets such as MIMIC III \cite{mimic3} and PhysioNet 2012 \cite{PhysioNet}, though public, are raw and require meticulous preprocessing to address issues like missing values or anomalies, adding another layer of variability. 

% Evaluation
\paragraph{L3: Ambiguous evaluation measures hinder a uniform and fair comparison} 
% Selection for TSG
Evaluating the generated time series hinges on three foundational principles: diversity, fidelity, and usefulness \cite{timegan}. 
Diversity gauges how closely the generated series distribution mirrors the original; fidelity examines the similarity between generated and real series; and usefulness assesses the generated series' practical utility in predictive tasks. 
The choice of evaluation measures crucially impacts method efficacy determination. 
Most studies \cite{timegan, psa-gan, timevae}, however, choose just a subset of these measures, introducing biases in performance evaluation. % We delve deeper into these measures in \sec{\ref{sec:benchmark:eval}}.
% Selection of downstream tasks.
In addition to quantitative evaluation for TSG, many methods \cite{rcgan, pategan} utilize downstream tasks, like classification and forecasting within the ``Train on Synthetic, Test on Real'' (TSTR) scheme, to showcase utility. 
These tasks, though popular, come with inherent challenges. For example, some datasets lack classification labels, and short sequences may be unsuitable for forecasting.
To counter this, alternative unsupervised approaches \cite{rcgan, ls4, som-vae} like interpolation and clustering have emerged, eliminating the need for external labels. 
However, they may introduce biases due to their dependency on post-hoc models in time series data \cite{gtgan, timevqvae}.

% why a benchmark is needed
To address these limitations, a comprehensive benchmark for TSG is crucial. 
It is worth noting that the time series community has introduced a plethora of benchmarks and surveys in areas like databases \cite{keogh2002need, hao2021ts}, classification \cite{ismail2019deep, uea, ucr}, forecasting \cite{makridakis2018m4, bauer2021libra, makridakis2021m5}, clustering \cite{rani2012recent, javed2020benchmark}, and anomaly detection \cite{lai2021revisiting, exathlon, tsb-uad, vus}. 
These benchmarks have been pivotal in driving advancements in their respective fields.
%%%
Nevertheless, there is a noticeable gap when it comes to the field of TSG despite its growing significance.
Some notable contributions include \citet{yan2022multifaceted}'s framework for benchmarking electronic health record generation and \citet{tsgm}'s open-source tool aimed at enhancing time series and fostering the use of generative models. 
A recent study \cite{eval_ts_tsg} also provides an overview of prevalent evaluation measures for TSG, complemented with an evaluation pipeline. 
Yet, to our knowledge, a holistic survey or benchmark dedicated to TSG is still absent.

\begin{figure}[ht]%
\centering%
\captionsetup{skip=0.75em,belowskip=0em}%
\includegraphics[width=0.99\columnwidth]{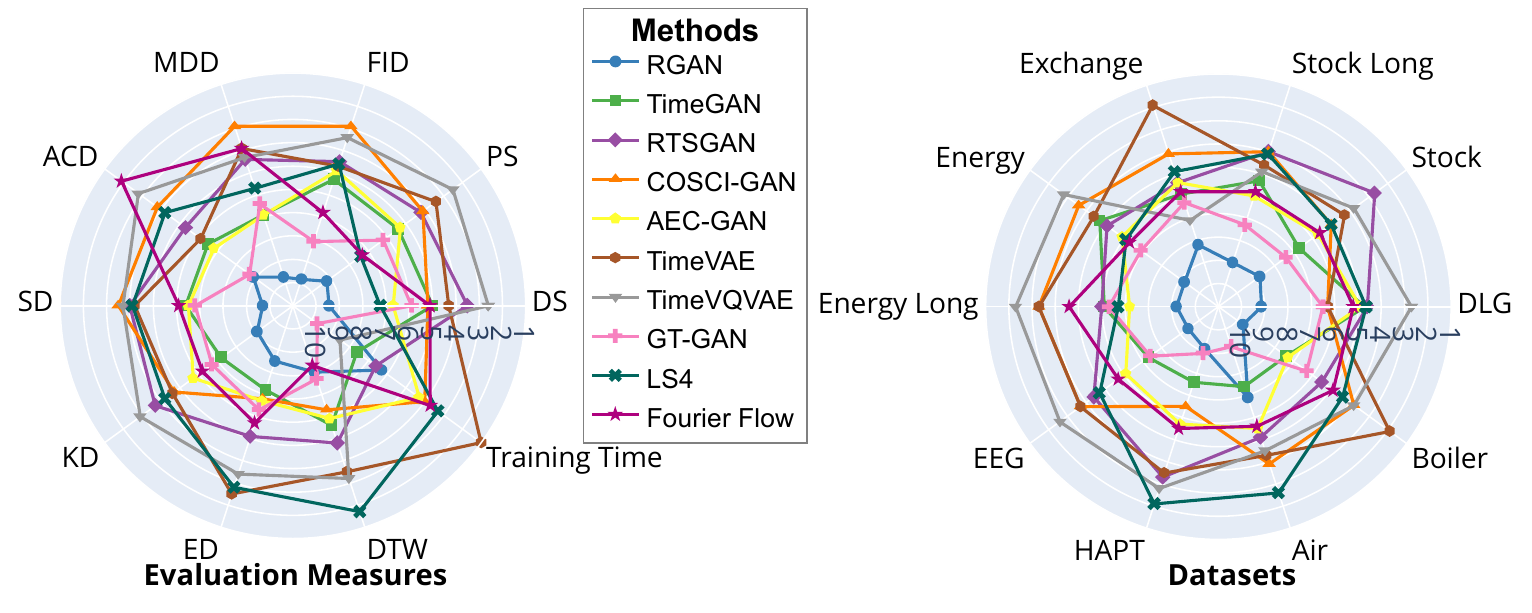}%
\caption{Method ranking across ten evaluation measures and ten datasets.}%
\label{fig:rank_all}%
\end{figure}%

\subsection{Our Contributions}
\label{sec:intro:contributions}

In light of these limitations and the evident gap for a dedicated TSG standard, we introduce \textsf{TSGBench}, an open-sourced benchmark designed to standardize comparative assessments of emerging methodologies.
It offers robust evaluations and in-depth analyses of principle methods, establishing a foundational benchmark for future investigations.
The key contributions of \textsf{TSGBench} (\textbf{C1--C4}) are highlighted as follows:

% taxonomy
\paragraph{C1: We provide a new taxonomy of various TSG methods based on three foundational generative models}
Regarding \textbf{L1}, we establish a taxonomy for systematic comparison and categorization (\sec{\ref{sec:taxonomy}}). 
Specifically, we categorize TSG methods based on three foundational generative models, offering a new perspective for understanding existing research.
On this basis, we compare and analyze the model they used, make connections if different methods use similar techniques, and elaborate upon the inheritance and improvement of relative methods. % (Figure \ref{fig:model} and Table \ref{tab:methods}). 

% public datasets
\paragraph{C2: We introduce a standard pipeline for dataset selection and preprocessing}
In terms of \textbf{L2}, we incorporate ten publicly available, real-world datasets from various application domains into \textsf{TSGBench} to enhance reproducibility and mitigate biases or oversimplification in our evaluations.
These datasets vary in terms of training sample size, sequence length, and time series dimensions, thereby contributing to a more comprehensive analysis.
Moreover, we introduce a standardized pipeline for preprocessing raw time series datasets for TSG (\sec{\ref{sec:benchmark:datasets}}). 

% evaluation suite and generalization test
\paragraph{C3: We design a suite of diverse yet robust measures to make a comprehensive assessment of the TSG}
For \textbf{L3}, we provide a suite of twelve evaluation measures tailored for TSG.
This suite encompasses five facets, i.e., model-based, feature-based, distance-based measures, training efficiency, and visualization. This allows for a standardized yet thorough comparison (\sec{\ref{sec:benchmark:eval}}).
%%%
Further, we introduce a novel generalization test based on Domain Adaptation (DA) to evaluate the generalization capabilities of TSG methods.
Notably, we define three scenarios (i.e., single DA, cross DA, and reference DA) that align with real-world applications (\sec{\ref{sec:benchmark:da}}).

% experiments
\paragraph{C4: We conduct systematic evaluations for TSG methods}
As part of the benchmark development, we present \textsf{TSGBench} in detail (\sec{\ref{sec:setup}}) and have conducted preliminary yet systematic experiments to assess the performance of ten representative TSG methods (\sec{\ref{sec:results}}).
The results demonstrate the ability of \textsf{TSGBench} to assess the efficacy of various TSG methods, illuminating their strengths and weaknesses from different perspectives. 
The generalization test using DA sheds light on domain shift scenarios in TSG. In addition, we provide statistical analysis of method ranking, examining the consistency of different techniques across ten evaluation measures and ten datasets (Figure \ref{fig:rank_all}). 
Our findings underscore the reliability and efficacy of \textsf{TSGBench} in evaluating TSG methods.

%% file: 04_preliminaries.tex
\section{Preliminaries}
\label{sec:prelim}

\subsection{Problem Definition}
\label{sec:prelim:definition}

% definition of time series
Suppose that a time series $\bm{T}$ with $N$ ($N \geq 1$) individual series of length $L$ is denoted as a matrix, i.e., $\bm{T}=(\bm{s}_1,\cdots,\bm{s}_N)^{\mathsf{T}}$, where each individual series $\bm{s}_i$ can be represented as an $L$-dimensional vector, i.e., $\bm{s}_i = (x_{i,1}, \cdots, x_{i,L})$, and each $x_{i,j}$ corresponds to a single time point $t_j$ of $\bm{s}_i$. 
% definition of TSG
We denote $p(\bm{s}_1,\cdots,\bm{s}_N)$ as the real distribution of a given time series $\bm{T}$. The goal of Time Series Generation (TSG) is to create a synthetic time series $\bm{T}^{gen} = (\bm{s}^{gen}_1,\cdots,\bm{s}^{gen}_N)$ such that its distribution $q(\bm{s}^{gen}_1,\cdots,\bm{s}^{gen}_N)$ is similar to $p(\bm{s}_1,\cdots,\bm{s}_N)$, and $\bm{T}^{gen}$ and $\bm{T}$ exhibit consistent statistical properties and patterns.
%%%
Table \ref{tab:notations} summarizes the frequently used notations in this work.

\subsection{Scope Illustration}
\label{sec:prelim:scope}

To ensure our initial benchmark is both focused and comprehensive, we employ specific constraints for \textsf{TSGBench}.

\paragraph{Scope of Methods}
%%% only consider general-purpose TSG methods. 
We restrict our attention to generation methods designed for general-purpose time series. 
Although certain methods showcase efficacy in specific domains, they lack the flexibility and adaptability needed for broader applications. 
Thus, these specialized methods are excluded from our evaluation.
%%% do not fine-tune each method
Moreover, while fine-tuning hyperparameters could yield superior results for each method, we opted not to engage in such optimization. 
This decision was made to maintain consistency and fairness in comparisons and to adhere to the resource and time constraints inherent in a comprehensive benchmarking study.

\paragraph{Scope of Datasets}
%%% diverse 
To make a holistic evaluation of different generation methods, we select public time series datasets that are both diverse and representative across application domains. 
%%% purely time series as the input 
Moreover, as we target the generation methods, we consider purely time series as the input. 
Thus, edge cases, such as those involving the input of a missing data-position matrix or the causality determined by causal graphs, are not incorporated.

\paragraph{Scope of Evaluation Measures}
We mainly adhere to the widely accepted ``Train on Synthetic, Test on Real'' (TSTR) scheme \cite{rcgan, pategan, timegan, rtsgan, imitation, tst-gan, gtgan, crvae, ls4}. It assesses the synthetic time series' relevance for real-world applications. 
While alternative schemes like ``Train on Real, Test on Synthetic'' (TRTS) \cite{cgan, cosci-gan} do exist, their likeness to TSTR and infrequent use render them non-essential for our purposes.
%%% exclude some measures such as cosine similarity
Our evaluation suite, illustrated in Figure \ref{fig:eval_heatmap}, is meticulously crafted using well-recognized measures from TSG literature. 
We have omitted measures specific to some TSG studies, like cosine similarity \cite{tts-gan}, due to their limited prevalence for extensive benchmarking.
%%% no more downstream tasks
Additionally, \textsf{TSGBench} does not use specific downstream tasks as separate evaluation metrics. 
This is because the included measures, notably discriminative and predictive scores, cover the main goals of time series downstream tasks, such as classification and forecasting. Furthermore, some tasks, like anomaly detection, require extra data and ground truth, which would complicate the evaluation and surpass the designated scope of \textsf{TSGBench}.
% due to two main reasons: 
% (1) The measures incorporated in \textsf{TSGBench}, notably the discriminative and predictive scores, already encompass key objectives of time series downstream tasks like classification and forecasting. 
% (2) Certain tasks (e.g., anomaly detection) demand the inclusion of supplementary data and corresponding ground truth, complicating the evaluation and surpassing the designated scope of \textsf{TSGBench}.

\input{tables/notations}

%% file: tables/notations.tex
\begin{table}[t]
\centering
\captionsetup{skip=0.75em}
\small
\caption{List of frequently used notations.}
\label{tab:notations}
\begin{tabular}{p{0.098\textwidth}p{0.34\textwidth}} \toprule
  \textbf{Symbol} & \textbf{Description} \\ \midrule
  $\bm{T}$, $N$ & A time series $\bm{T}$ with $N$ individual series \\
  $\bm{s}_i$, $L$ & An individual series $\bm{s}_i$ of length $L$ \\
  $l$ & The sequence length for partitioning \\ 
  $R$ & The number of sub-matrices, where $R=L-l+1$ \\
  $\bm{T}_r$ & A sub-matrix $\bm{T}_r$ with $N$ individual series of length $l$ \\ %  ($1 \leq r \leq R$)
  $\bm{T}_s^{tr}$, $\bm{T}_s^{te}$  & The training (or historical) data and the test data from a source time series \\
  $\bm{T}_t^{his}$,$\bm{T}_t^{gen}$,$\bm{T}_t^{gt}$ & The historical, generated, and ground truth data from a target time series \\
  \bottomrule 
\end{tabular}
\end{table}

% # series --> N
% # Time points --> L
% # Samples --> R
% seq_len --> l
% shape of (L, N) --> shape of (R, l, N)

%% file: 05_taxonomy.tex
\section{TSG Overview}
\label{sec:taxonomy}

This section provides a taxonomy and comprehensive analysis of TSG methods. We first explore three foundational generative models: GAN, VAE, and Flow-based models (\sec{\ref{sec:taxonomy:model}}). We then delve into notable TSG methods stemming from these models (\sec{\ref{sec:taxonomy:methods}}). 

\begin{figure*}[t]
\centering
\captionsetup{skip=0.0em}%
\subfigure[GAN-based models.]{%
  \label{fig:model:gan}%
  \includegraphics[width=0.325\textwidth]{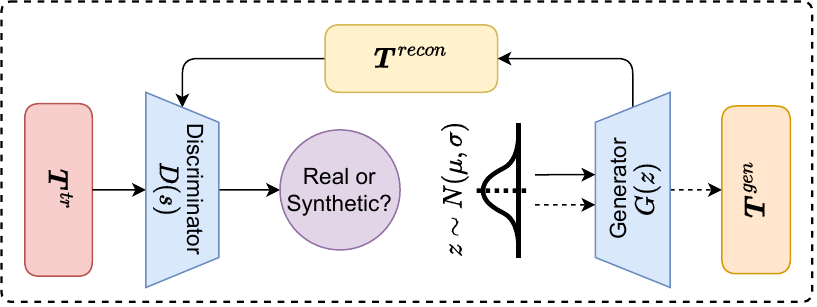}}%
\hspace{0.1em}
\subfigure[VAE-based models.]{%
  \label{fig:model:vae}%
  \includegraphics[width=0.325\textwidth]{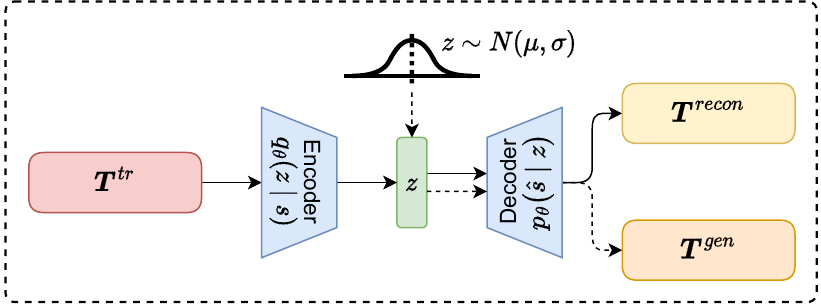}}%
\hspace{0.1em}
\subfigure[Flow-based models.]{%
  \label{fig:model:flow}%
  \includegraphics[width=0.325\textwidth]{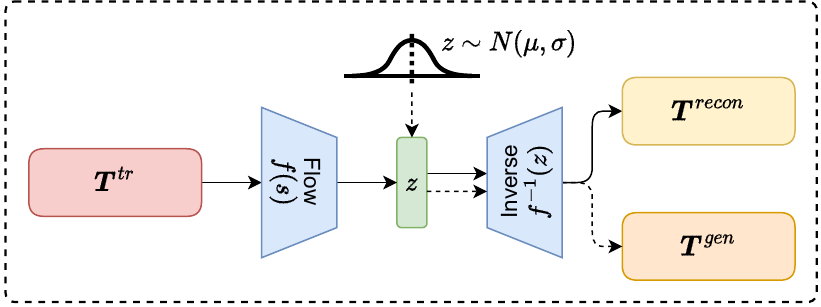}}%
\caption{Three foundational generative models in popular TSG methods (solid arrow: training, dashed arrow: generation).}
\label{fig:model}
\end{figure*}

\subsection{Generative Models for TSG}
\label{sec:taxonomy:model}

Generative models are designed to learn the intricate patterns and temporal dependencies in time series datasets, allowing for the generation of new time series that reflect the original data's statistical properties.  
We next describe three foundational generative models for TSG, with their architectural nuances visualized in Figure \ref{fig:model}. 

\paragraph{Generative Adversarial Networks (GANs)}
%%% architecture of GAN
A typical GAN model \cite{goodfellow2014generative}, as depicted in Figure \ref{fig:model:gan}, comprises a generator $G$ and a discriminator $D$. 
During training, $G$ produces a synthetic time series that $D$ attempts to distinguish from real series. 
This adversarial cycle continues until $D$ cannot reliably differentiate between the two. 
For generation, only $G$ is used to create a new time series.

%%% adapt GANs to TSG
Numerous algorithms employ GAN-based models for TSG.
They often incorporate specialized neural networks like RNN, LSTM, and Transformer to capture the sequential and temporal intricacies of time series data \cite{crnngan, rcgan, tts-gan, tst-gan, timegan, rtsgan}. 
%%%
Some models utilize novel metrics or loss functions for better alignment with specific temporal patterns \cite{sig-gan, c-sig-gan}. 
Others enhance traditional GANs with additional modules, such as extra discriminators, classification layers, error correction, and data augmentation to generate time series with specific temporal attributes \cite{aec-gan, cosci-gan, psa-gan}.
%%%
While GAN-based models are effective in generating time series, they can be challenging to train and are often resource and time-intensive \cite{gtgan}.

\paragraph{Variational AutoEncoders (VAEs)}
%%% architecture of VAEs
A standard VAE model \cite{kingma2013auto, kingma2014semi}, as shown in Figure \ref{fig:model:vae}, contains an encoder and a decoder. 
The encoder $q_{\theta}(\bm{z} | \bm{s})$ transforms input time series $\bm{s}$ into a latent representation $\bm{z}$, capturing essential features and generating parameters (e.g., mean $\mu$ and variance $\sigma$) that model the inherent uncertainty and variability of the time series.
The decoder $p_{\theta}(\hat{\bm{s}} | \bm{z})$ reconstructs time series $\hat{\bm{s}}$ from $\bm{z}$, preserving the model's capacity to regenerate the temporal patterns. 
%%% training and generation
The training phase focuses on minimizing reconstruction loss and the divergence between the learned and a prior standard Gaussian distribution. 
For generation, the decoder draws from the latent space to create a synthetic time series.

While there are fewer studies on VAE-based methods for TSG compared to GANs, they effectively leverage variational inference to capture the complex temporal aspects of time series data \cite{timevae, crvae, timevqvae}. 
%%% advantages 
The benefits of VAE models lie in their interpretability and training efficiency. Additionally, the structured and informative latent representations from VAEs can be used for other time series tasks, including representation learning \cite{som-vae} and imputation \cite{gp-vae}.

\paragraph{Flow-based Models}
% https://lilianweng.github.io/posts/2018-10-13-flow-models/
GANs and VAEs do not directly model the probability density function of time series due to the computational challenge of covering all possible latent representation $\bm{z}$ values. 
Flow-based models \cite{nice, realnvp, glow}, as shown in Figure \ref{fig:model:flow}, address this by using invertible transformations $f$'s to explicitly learn data distributions.
%%%
They have been increasingly employed in TSG, often utilizing explicit likelihood models or Ordinary Differential Equations (ODEs) \cite{NODE, ode-rnn, nsde, gtgan, fourier}. 
Their architectures, typically featuring coupling layers, allow for a computable Jacobian determinant and reversibility. 
This is further enhanced by specific transformation techniques crucial for modeling complex data distributions.

\subsection{Taxonomy of TSG Methods}
\label{sec:taxonomy:methods}

We detail ten representative TSG methods (\textbf{A1--A10}) grounded in the three foundational generative models. 
Table \ref{tab:methods} provides a summary of these methods, including their backbone models.

\input{tables/methods}

\paragraph{Pure GAN-based Methods}
Early efforts \cite{crnngan, rcgan, t-cgan} blended vanilla GAN architectures from image generation with neural networks like RNN and LSTM tailored for sequential data. Subsequent studies have focused on pioneering techniques to adapt to time series data and boost performance.
\begin{itemize}[nolistsep,leftmargin=20pt]
  \item \keypoint{A1: RGAN \cite{rcgan}.} RGAN is a pioneering work that utilizes the GANs for TSG based on RNNs. It is inspired by the maximum mean discrepancy \cite{MMD} and tries to measure the statistical difference between generated and practical time series.
  
  \item \keypoint{A2: TimeGAN \cite{timegan}.} 
  TimeGAN considers temporal dependencies within GANs, by simultaneously learning to encode features, generate representations, and iterate across time. It delivers advanced performance
  and has become a benchmark model for subsequent methods.
  
  \item \keypoint{A3: RTSGAN \cite{rtsgan}.} 
  RTSGAN combines an autoencoder into GANs and centers on generating time series with varying lengths and handling missing data.

  \item \keypoint{A4: COSCI-GAN \cite{cosci-gan}.} 
  COSCI-GAN is devised to explicitly model the complex dynamical patterns across every series, which favors channel/feature relationship preservation.
  
  \item \keypoint{A5: AEC-GAN \cite{aec-gan}.} 
  AEC-GAN aims to generate long time series with distribution shifts and bias amplification via an error correction module that corrects bias in previously generated data and introduces adversarial samples.
\end{itemize}

\paragraph{Pure VAE-based Methods}
VAE-based methods often exploit variational inference to capture the temporal features effectively. They are generally efficient and have the potential interpretability.
\begin{itemize}[nolistsep,leftmargin=20pt]
  \item \keypoint{A6: TimeVAE \cite{timevae}.} 
  TimeVAE extends VAEs to the general-purpose TSG. It builds on convolution and enhances interpretability through time series decomposition.
  
  \item \keypoint{A7: TimeVQVAE \cite{timevqvae}.} It employs the STFT to decompose input time series into low-frequency and high-frequency components. Then, it integrates Vector Quantization with VAEs \cite{vqvae} for enhanced modeling of these components, preserving both the general shape and specific details of the time series.
\end{itemize}

\paragraph{Mixed-Type Methods}
Recent TSG advancements have explored mixed-type methods, merging flow-based models with techniques like DFT or ODEs, or integrating them with GANs or VAEs.
\begin{itemize}[nolistsep,leftmargin=20pt]
  \item \keypoint{A8: Fourier Flows \cite{fourier}.} 
  It uses DFT \cite{oppenheim1999discrete} to analyze the time series in the frequency domain and applies a sequence of data-dependent spectral filters to learn their distributions.

  \item \keypoint{A9: GT-GAN \cite{gtgan}.} 
  GT-GAN is tailored for dealing with both regular and irregular time series, employing Continuous Time Flow Processes (CTFP) \cite{ctfp} for its generator and GRU-ODE for the discriminator;

  \item \keypoint{A10: LS4 \cite{ls4}.} 
  LS4 draws from deep-state space models and incorporates stochastic latent variables to enhance the model's capacity and leverage the training objectives from VAEs.  
\end{itemize}

%% file: tables/methods.tex
\begin{table}[t]
\centering
\captionsetup{skip=0.75em}
\caption{Summary of popular TSG methods with their backbone models and specialties (TS: Time Series).}
\label{tab:methods}%
\resizebox{\columnwidth}{!}{%
\begin{tabular}{cccc} \toprule
  \textbf{Time} & \textbf{Method} & \textbf{Model} & \textbf{Specialty} \\
  \midrule
  2016  & C-RNN-GAN \cite{crnngan} & GAN   & Music \\
  2017  & RGAN \cite{rcgan}  & GAN   & General (w/ Medical) TS \\
  2018  & T-CGAN \cite{t-cgan} & GAN   & Irregular TS \\
  2019  & WaveGAN \cite{wavegan} & GAN   & Audio \\
  2019  & TimeGAN \cite{timegan} & GAN   & General TS \\
  2020  & TSGAN \cite{cgan} & GAN   & General TS \\
  2020  & DoppelGANger \cite{DoppelGANger}  & GAN   & General TS \\
  2020  & SigCWGAN \cite{c-sig-gan}  & GAN   & Long Financial TS \\
  2020  & Quant GANs \cite{quant-gan}  & GAN   & Long Financial TS \\
  2020  & COT-GAN \cite{cot-gan} & GAN   & TS and Video \\
  2021  & Sig-WGAN \cite{sig-gan} & GAN   & Financial TS \\
  2021  & TimeGCI \cite{imitation} & GAN   & General TS \\
  2021  & RTSGAN \cite{rtsgan} & GAN   & General (w/ Incomplete) TS \\
  2022  & PSA-GAN \cite{psa-gan} & GAN   & General (w/ Forecasting) TS \\
  2022  & CEGEN \cite{cegen} & GAN   & General TS \\
  2022  & TTS-GAN \cite{tts-gan} & GAN   & General TS \\
  2022  & TsT-GAN \cite{tst-gan} & GAN   & General TS \\
  2022  & COSCI-GAN \cite{cosci-gan} & GAN   & General TS \\
  2023  & AEC-GAN \cite{aec-gan}  & GAN   & Long TS \\
  2023  & TT-AAE \cite{tt-aae} & GAN   & General TS \\
  \cmidrule{1-4} % \midrule
  2021  & TimeVAE \cite{timevae} & VAE   & General TS \\
  2023  & CRVAE \cite{crvae} & VAE   & Medical TS \& Causal Discovery \\
  2023 & TimeVQVAE \cite{timevqvae} & VAE & General TS     \\
  \cmidrule{1-4} % \midrule
  2018  & Neural ODE \cite{NODE} & ODE + RNN   & General TS \\
  2019  & ODE-RNN \cite{ode-rnn} & ODE + RNN   & Irregular TS \\
  2021  & Neural SDE \cite{nsde} & ODE + GAN & General TS \\
  2022  & GT-GAN \cite{gtgan} & ODE + GAN & General (w/ Irregular) TS \\
  2023  & LS4 \cite{ls4}   & ODE + VAE & General (w/ Forecasting) TS \\
  2020  & CTFP \cite{ctfp}  & Flow   & General TS \\
  2021  & Fourier Flow \cite{fourier} & Flow   & General TS \\
  % \cmidrule{1-4} % \midrule
  % 2016  & WaveNet \cite{wavenet} & CNN   & Speech \\
  2023  & TSGM \cite{sgm}   & SGM & General TS \\
  \bottomrule
\end{tabular}%
}
\end{table}%

%% file: 06_benchmark.tex
\section{TSGBench}
\label{sec:benchmark}

\begin{figure*}[t]%
\centering%
\captionsetup{skip=0.75em,belowskip=0em}%
\includegraphics[width=0.99\textwidth]{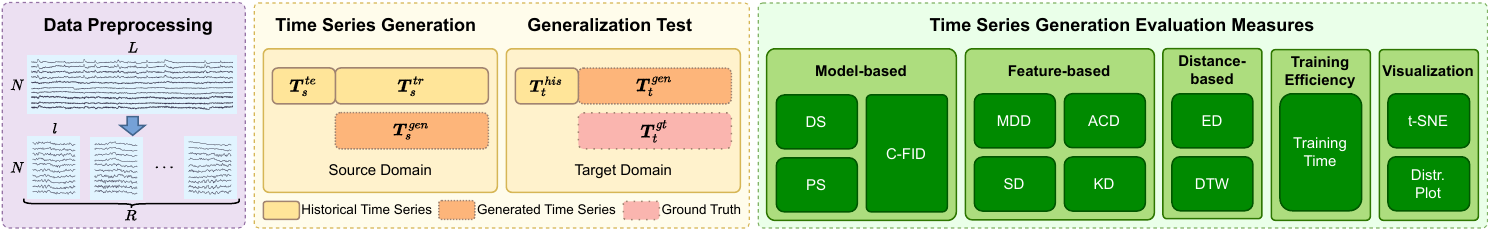}%
\caption{Overall architecture of \textsf{TSGBench}.}%
\label{fig:tsg-bench}%
\end{figure*}

To address the challenges and potential biases in dataset and evaluation selection, we have distilled the best practices from pertinent research and established \textsf{TSGBench}, a benchmark tailored for systematically assessing TSG methods.
Its architecture is visualized in Figure \ref{fig:tsg-bench}, which encompasses three key modules:
(1) a meticulous set of ten public, real-world time series datasets with a standardized preprocessing pipeline (\sec{\ref{sec:benchmark:datasets}});
(2) a comprehensive suite of twelve evaluation measures customized for TSG (\sec{\ref{sec:benchmark:eval}});
(3) an innovative generation test using Domain Adaption (DA) for TSG (\sec{\ref{sec:benchmark:da}}).

\subsection{Dataset Selection and Preprocessing}
\label{sec:benchmark:datasets}
\textbf{Dataset Selection.}
We use only publicly available, real-world datasets to ensure reproducibility and sidestep biases or oversimplification in our evaluations. 
Importantly, our aim is \emph{not} to amass an exhaustive dataset collection but to curate a diverse set that spans multiple domains and features varied data statistics and distributions. 
We identify and collect ten datasets (\textbf{D1--D10}) published over the past two decades. 
Table \ref{tab:dataset} summarizes their statistics. 
Below, we provide a brief description of each dataset.
\begin{itemize}[nolistsep,leftmargin=20pt]
  \item \keypoint{D1: Dodgers Loop Game (DLG) \cite{misc_dodgers_loop_sensor_157}.} 
  It consists of loop sensor data from the Glendale on-ramp for the 101 North freeway in Los Angeles.
  
  \item \keypoint{D2: Stock \cite{timegan}.} 
  It comprises daily historical Google stock data from 2004 to 2019, including volume and high, low, opening, closing, and adjusted closing prices. 
  
  \item \keypoint{D3: Stock Long \cite{timegan}.} 
  It is identical to the Stock dataset but with a sequence length of 125.

  \item \keypoint{D4: Exchange \cite{lai2018modeling}.} 
  It contains the daily exchange rates of eight countries (i.e., Australia, Britain, Canada, Switzerland, China, Japan, New Zealand, and Singapore) from 1990 to 2016.

  \item \keypoint{D5: Energy \cite{misc_appliances_energy_prediction_374}.} 
  It includes information on appliance's energy use in a low-energy building.

  \item \keypoint{D6: Energy Long \cite{misc_appliances_energy_prediction_374}}. 
  It is identical to the Energy dataset but with a sequence length of 125.
  
  \item \keypoint{D7: EEG \cite{eeg}.} 
  It is with the measurements derived from ElectroEncephaloGraphy (EEG) data captured by Emotiv EEG Neuroheadset. It helps to understand brainwave patterns, especially those under specific cognitive conditions or stimuli.
  
  \item \keypoint{D8: HAPT \cite{misc_human_activity_recognition_using_smartphones_240}.} 
  It comprises recordings of 30 subjects performing activities of daily living captured via waist-mounted smartphones with embedded inertial sensors.

  \item \keypoint{D9: Air \cite{zheng2015forecasting}.} 
  It has air quality, meteorological, and weather forecast data from 4 major Chinese cities: Beijing, Tianjin, Guangzhou, and Shenzhen from 2014/05/01 to 2015/04/30.

  \item \keypoint{D10: Boiler \cite{sasa}.} 
  It collects sensor data from three boilers from 2014/03/24 to 2016/11/30 to monitor the operating states.
\end{itemize}

\input{tables/datasets}

\paragraph{Preprocessing Pipeline}
While these ten datasets are publicly accessible, few TSG methods have evaluated a majority of them. 
To circumvent these issues, we introduce a standardized pipeline for preprocessing the raw time series datasets tailored for TSG. 

To generate time series in a brief span while preserving meaningful structures, we first follow \cite{timegan, imitation} and segment the long time series $\bm{T}$ into shorter sub-matrices $\{\bm{T}_1,\bm{T}_2,\bm{T}_3,\cdots\}$. 
With a specified sequence length $l$ and a stride of 1, we convert $\bm{T}$ into $R$ overlapping sub-matrices $\{\bm{T}_r\}_{1 \leq r \leq R}$, where $R = L-l+1$ and each $\bm{T}_r$ has the same $l$.
%%% seq length
To determine the value of $l$, we employ autocorrelation functions \cite{parzen1963spectral}, ensuring that each $\bm{T}_r$ encompasses at least one time series period. 
%%% sliding window
The time series is then shuffled to approximate an i.i.d. sample distribution \cite{timegan}. 
%%% train/test split
To assess the generalization capability of TSG methods, we divide the data into training and testing sets in a 9:1 ratio, allocating a larger portion for training and evaluation as is common in TSG methodology. 
%%% standardize to [0,1]
Additionally, we normalize the dataset to the range of $[0,1]$ to enhance efficiency and numerical stability, resulting in a dataset shape of $(R, l, N)$.

\subsection{Evaluation Measure Suite}
\label{sec:benchmark:eval}

\begin{figure}[t]
\centering%
\captionsetup{skip=0.75em,belowskip=0em}%
\includegraphics[width=0.99\columnwidth]{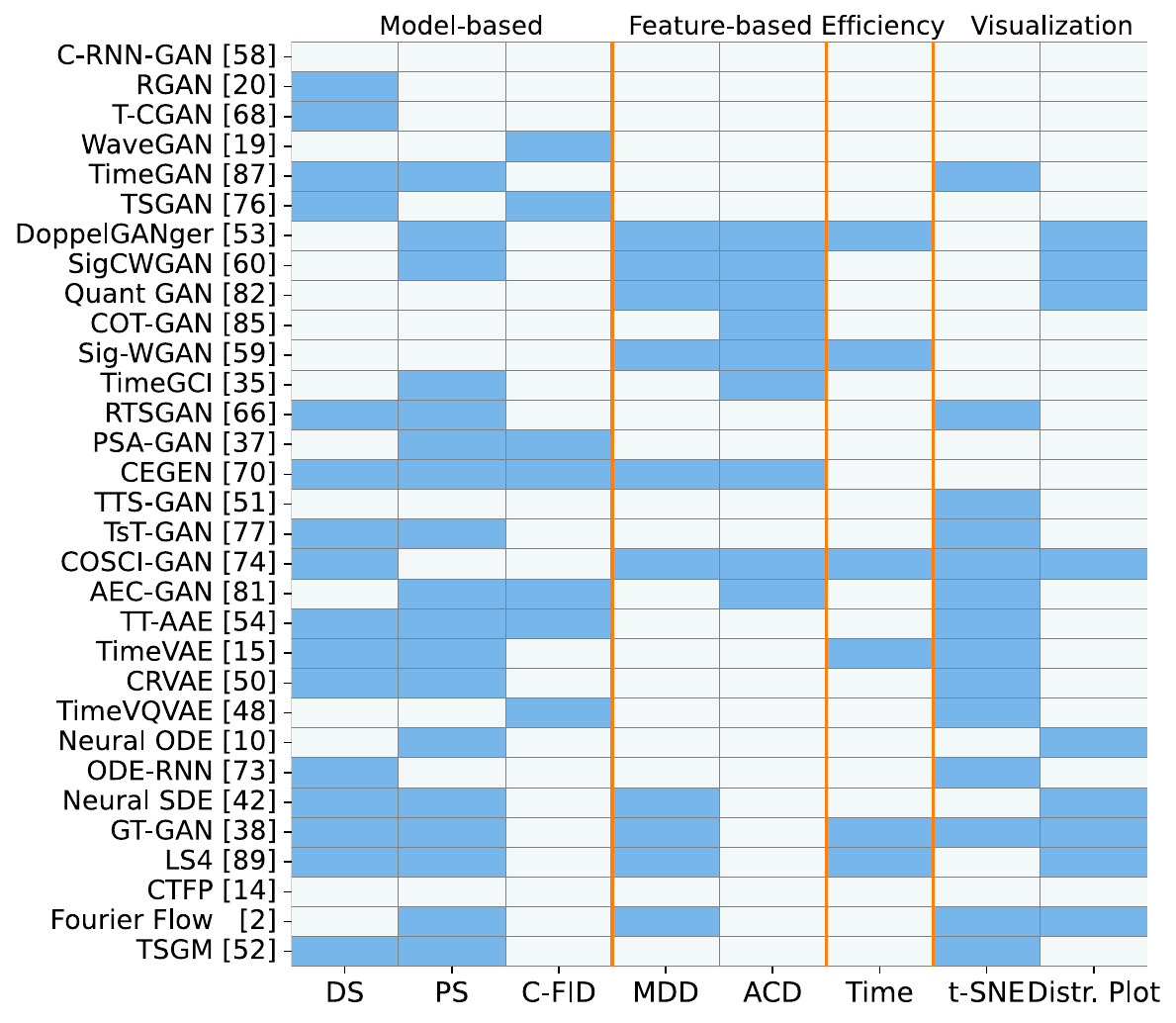}%
\caption{Summary of evaluations in popular TSG methods.}%
\label{fig:eval_heatmap}%
\end{figure}

% Description and taxonomy of current measures
Dozens of measures exist to gauge the quality of TSG methods, which typically adhere to principles like diversity, fidelity, and usefulness, as outlined in \sec{\ref{sec:intro:motivation}}.
We next offer a suite of twelve prevalent measures (\textbf{M1--M12}), complemented by in-depth descriptions. A summary of these measures used in TSG is depicted in Figure \ref{fig:eval_heatmap}.

\paragraph{Model-based Measures}
These measures predominantly adhere to the TSTR scheme \cite{rcgan, pategan}. This scheme involves using the synthetically generated series to train a post-hoc neural network, which is subsequently tested on the original time series. 

\begin{itemize}[nolistsep,leftmargin=20pt]
  \item \keypoint{M1: Discriminative Score (DS) \cite{timegan}.} 
  This measure employs a post-hoc time-series classification model with 2-layer GRUs or LSTMs to differentiate between original and generated series \cite{timegan}. 
  Each original series is labeled as \emph{real}, while the generated series is labeled \emph{synthetic}. 
  Using these labels, an RNN classifier is trained. The classification error on a test set quantifies the generation model's fidelity.
  
  \item \keypoint{M2: Predictive Score (PS) \cite{timegan}.} 
  It involves training a post-hoc time series prediction model on synthetic data \cite{timegan}. 
  Using GRUs or LSTMs, the model predicts either the temporal vectors of each input series for the upcoming steps \cite{timegan, imitation} or the entire vector \cite{gtgan}. 
  The model's performance is then evaluated on the original dataset using the mean absolute error.
  
  \item \keypoint{M3: Contextual-FID (C-FID) \cite{psa-gan}.} 
  It extends the concept of Frechet Inception Distance (FID) \cite{fid} from image generation to TSG. It quantifies how well the synthetic time series conforms to the local context of the time series. Using the time series embeddings from \cite{ts2vec}, it learns embeddings that seamlessly blend with the local context.
\end{itemize}

\paragraph{Feature-based Measures}
These measures are designed to capture inter-series correlations and temporal dependencies, assessing how well the generated time series preserves original characteristics. 
A key advantage of feature-based measures is their capacity to yield clear and deterministic results, providing an unambiguous assessment of the quality of generated time series.
\begin{itemize}[nolistsep,leftmargin=20pt]
  \item \keypoint{M4: Marginal Distribution Difference (MDD) \cite{sig-gan}.}
  This measure computes an empirical histogram for each dimension and time step in the generated series, using the bin centers and widths from the original series. 
  It then calculates the average absolute difference between this histogram and that of the original series across bins, assessing how closely the distributions of the original and generated series align.

  \item \keypoint{M5: AutoCorrelation Difference (ACD) \cite{lai2018modeling}.} 
  It computes the autocorrelation of both the original and generated time series, then determines their difference \cite{parzen1963spectral, lai2018modeling}.
  By contrasting the autocorrelations, we could evaluate how well dependencies are maintained in the generated time series.

  \item \keypoint{M6: Skewness Difference (SD).} 
  Beyond ACF, we also consider the statistical measures \cite{aec-gan, eval_ts_tsg}. Skewness is vital for the marginal distribution of a time series, quantifying its distribution asymmetry. 
  Given the mean (standard deviation) of the train time series $\bm{T}_s^{tr}$ as $\bm{\mu}_s^{tr}$ ($\bm{\sigma}_s^{tr}$) and the generated time series $\bm{T}_s^{gen}$ as $\bm{\mu}_s^{gen}$ ($\bm{\sigma}_s^{gen}$), 
  we evaluate the fidelity of $\bm{T}_s^{gen}$ by computing the skewness difference between them as:
  \begin{equation}\label{eqn:skewness}
    SD = \abso{\frac{\mathbb{E}[(\bm{T}_s^{gen}-\bm{\mu}_s^{gen})^3]}{{\bm{\sigma}_s^{gen}}^3} - \frac{\mathbb{E}[(\bm{T}_s^{tr}-\bm{\mu}_s^{tr})^3]}{{\bm{\sigma}_s^{tr}}^3}}.
  \end{equation}

  \item \keypoint{M7: Kurtosis Difference (KD).} 
  Like skewness, kurtosis assesses the tail behavior of a distribution, revealing extreme deviations from the mean. 
  Using notations from Equation \ref{eqn:skewness}, the kurtosis difference between $\bm{T}_s^{tr}$ and $\bm{T}_s^{gen}$ is calculated as:
  \begin{equation}\label{eqn:kurtosis}
    KD = \abso{\frac{\mathbb{E}[(\bm{T}_s^{gen}-\bm{\mu}_s^{gen})^4]}{{\bm{\sigma}_s^{gen}}^4} - \frac{\mathbb{E}[(\bm{T}_s^{tr}-\bm{\mu}_s^{tr})^4]}{{\bm{\sigma}_s^{tr}}^4}}.
  \end{equation}
\end{itemize}

\paragraph{Training Efficiency}
Training efficiency is crucial, particularly in cases that demand rapid TSG methods or where computational resources are scarce. 
However, only a few studies, such as \cite{timevae, gtgan}, have been employed for evaluation in this context.
\begin{itemize}[nolistsep,leftmargin=20pt]
  \item \keypoint{M8: Training Time.} 
  It refers to the wall clock time for training a TSG method. It is a vital measure for evaluating and deploying TSG methods due to economic considerations.
  % It refers to the wall clock time it takes for a TSG method to learn from a dataset during the training phase. 
  % It is a vital measure in the evaluation and deployment of TSG methods, both for practical and economic reasons.
\end{itemize}

\paragraph{Visualization}
Visualization offers an intuitive and visually interpretive perspective to directly compare and contrast the structures and patterns between the original and generated time series.
\begin{itemize}[nolistsep,leftmargin=20pt]
  \item \keypoint{M9: t-SNE \cite{tsne}.} 
  It is a prevalent technique for succinctly visualizing the distribution of generated time series compared to the original one within a two-dimensional space.
  
  \item \keypoint{M10: Distribution Plot \cite{gtgan}.} 
  It illuminates the difference between the input and generated time series in terms of density, spread, and central tendency to show how the generated time series closely mirrors the original's statistics.
\end{itemize}

\paragraph{Challenges in DS and PS}
As depicted in Figure \ref{fig:eval_heatmap}, DS and PS are frequently employed for evaluating TSG methods. However, they can yield unreliable evaluations for several reasons:
\begin{enumerate}[nolistsep,leftmargin=25pt]
  %%% random & unstable => time-consuming
  \item DS and PS rely on the TSTR scheme, necessitating post-hoc model training. The inherent randomness in deep models, such as weight initialization, can cause inconsistent evaluations \cite{gtgan, timevqvae}. While repeated training and averaging of results can mitigate this issue, it is time-intensive.
  % DS and PS rely on the TSTR scheme, which requires post-hoc model training. Deep models often introduce randomness, like weight initialization, that can cause inconsistent evaluations \cite{gtgan, timevqvae}. 
  % Although repeating training and averaging results can help, this process is time-consuming.
  
  %%% different architectures & configurations
  \item These measures often hinge on specific network architectures and configurations, which vary across studies \cite{timegan, psa-gan}, complicating the comparisons of different TSG methods.

  %%% sensitive to #samples
  \item They are sensitive to dataset sizes. With smaller datasets, they may not converge effectively due to insufficient training data, compromising their reliability \cite{timevae, timevqvae}.

  %%% short sequence length is not suitable for downstream tasks
  \item The sequence length for TSG is typically short, often just 24-time points, as seen in datasets like Stock and Energy \cite{timegan}. 
  They may not contain enough periodic information to accurately assess the TSG efficacy in real-world scenarios.
\end{enumerate}

In summary, while DS and PS offer specific insights into the quality of generated time series, their lack of robustness, time-intensive nature, and difficulty adapting to varying dataset sizes and sequence lengths make them less suitable for a comprehensive and reliable evaluation of TSG methods.

\paragraph{Distance-based Measures}
To mitigate the challenges associated with DS and PS, we propose the incorporation of two distance-based measures to provide an efficient, deterministic evaluation.
\begin{itemize}[nolistsep,leftmargin=20pt]
  \item \keypoint{M11: Euclidean Distance (ED).} 
  For each original series $\bm{s}^{tr} = (x_1,\cdots,x_l)$ and its generated series $\bm{s}^{gen} = (y_1,\cdots,y_l)$, $ED = \sqrt{\textstyle \sum_{i=1}^{l} (x_{i} - y_{i})^2}$.
  We take the mean of ED for all series and all samples.
  Given that the input time series has been preprocessed to fit within the range of $[0,1]$, ED deterministically assesses the similarity between $\bm{s}^{gen}$ and $\bm{s}^{tr}$. It provides a value-wise comparison between the time series.

  \item \keypoint{M12: Dynamic Time Warping (DTW) \cite{dtw}.}
  Given that ED overlooks alignment, we include DTW to capture the optimal alignment between series regardless of their pace or timing. 
  The alignment facilitated by DTW offers insights into the predictive quality of the generated series. Moreover, as shown by \cite{shokoohi2017generalizing}, multi-dimensional DTW \cite{meert_wannes_2020_7158824} can enhance downstream classification tasks, serving as a discriminative measure.
\end{itemize}

Using ED and DTW, we can efficiently and effectively evaluate the quality of generated time series, offering streamlined alternatives to DS and PS with similar evaluation goals.

\subsection{Generalization Test}
\label{sec:benchmark:da}

\textbf{Motivation.}
The domain shift problem is a significant concern in the field of time series analysis \cite{sasa, adatime, Datsing}. 
For the task of TSG, most methods usually require large datasets, and many GAN-based methods (e.g., TimeGAN \cite{timegan} and GT-GAN \cite{gtgan}) are time-intensive to train and fine-tune \cite{timevae}.
Nevertheless, many applications struggle to quickly accumulate sufficient data. Their efficacy may suffer in data-limited situations. Thus, evaluating the generalization capabilities of these methods on small data becomes crucial.

\paragraph{Domain Adaptation (DA)}
We introduce a novel generalization test using DA \cite{alaa2022faithful} to assess the generalization capabilities of TSG methods on small datasets. 
While DA has been widely applied in other time series tasks like classification and forecasting \cite{sasa, CoDATS, Datsing}, its application in TSG remains underexplored. 
This is because typical DA tasks necessitate labels (such as class labels for classification or future values for forecasting) and aim to minimize the distribution shift between source and target domains. % under the assumption of a shared label space.

We follow the conventions in DA tasks for time series data \cite{adatime, sasa, CoDATS, Datsing}, examining examples that range from monitoring patient mortality \cite{misc_human_activity_recognition_using_smartphones_240} to recording climate and air quality across cities \cite{zheng2015forecasting}, and assessing different machines in a factory \cite{sasa}. 
The commonality across these examples lies in the time-dependent nature of the data, while the differences come from the unique characteristics inherent to each scenario.
To illustrate, we first present a motivating example before offering a formal definition.

\begin{example}
\label{exp:da}
  In a factory, a machine (source domain) equipped with sensors generates operational data as a source time series $\bm{T}_s$.
  A TSG model $G$ is trained on this machine's historical data $\bm{T}_s^{tr}$ and tested on its new data $\bm{T}_s^{te}$.
  %%%
  When a new machine (target domain) with identical sensors is installed, TSG models are used to synthesize sensor readings due to limited data availability. 
  %%% 
  After collecting a brief historical time series $\bm{T}_t^{his}$ from this new machine, the goal is to generate a synthetic time series $\bm{T}_t^{gen}$ reflecting its expected performance and patterns. 
  %%%
  An effective TSG model should adapt to this new machine, producing realistic synthetic data.
  \hfill $\triangle$ \par 
\end{example}

In Example \ref{exp:da}, the time series available are $\bm{T}_s^{tr}$, $\bm{T}_s^{te}$, and $\bm{T}_t^{his}$. 
To evaluate the generated time series $\bm{T}_t^{gen}$, we use a comprehensive time series  $\bm{T}_t^{gt}$ from the target domain as the ground truth.
%%%
Since we aim to benchmark TSG methods, we do not explore scenarios like modifying the model's architectures or hyperparameter tuning. Accordingly, we focus on the following three DA cases.
\begin{definition}[Single DA]
\label{def:da:single}
  A TSG model $G$ is trained using the time series $\bm{T}_s^{tr}$ from the source domain. 
  It then generates a new time series $\bm{T}_t^{gen}$ in the target domain and evaluates the performance against $\bm{T}_t^{gt}$.
\end{definition}

\begin{definition}[Cross DA]
\label{def:da:cross}
  A TSG model $G$ is trained using the time series $\bm{T}_s^{tr}$ from the source domain, alongside a small subset of time series from the target domain $\bm{T}_t^{his}$. 
  It then generates a new time series $\bm{T}_t^{gen}$ in the target domain and assesses the performance against $\bm{T}_t^{gt}$.
\end{definition}

\begin{definition}[Reference DA]
\label{def:da:baseline}
  A TSG model $G$ is trained solely using a small subset of time series from the target domain $\bm{T}_t^{his}$. 
  It then generates a new time series $\bm{T}_t^{gen}$ in the target domain and gauges the performance against $\bm{T}_t^{gt}$.
\end{definition}

\paragraph{Datasets and Evaluations}
We extend three datasets HAPT \cite{misc_human_activity_recognition_using_smartphones_240}, Air \cite{zheng2015forecasting}, and Boiler \cite{sasa} for generalization testing, as they include domain information in line with conventions in time series DA tasks \cite{adatime, sasa}. 
Specifically, our dataset configurations are as follows:
\begin{itemize}[nolistsep,leftmargin=20pt]
  \item \keypoint{HAPT \cite{misc_human_activity_recognition_using_smartphones_240}.} 
  The user is treated as the domain attribute. 
  We randomly select User 14 as the source domain and Users 0, 23, 18, 52, and 20 as the target domains. 
  Our evaluation targets the time series for `walking' as it provides a more concentrated and comparable analysis of each user's walking pattern.

  \item \keypoint{Air \cite{zheng2015forecasting}.} 
  The city serves as the domain attribute. 
  We randomly chose Tianjin (TJ) as the source domain and selected Beijing (BJ), Guangzhou (GZ), and Shenzhen (SZ) as the target domains. 
  This setup enables us to investigate variations in air quality patterns across different urban settings.

  \item \keypoint{Boiler \cite{sasa}.} 
  We employ the boiler machine as the domain attribute. Boiler 1 is randomly chosen as the source domain, with Boilers 2 and 3 serving as the target domains. 
  This configuration allows us to examine operational variations and similarities across different boiler units.
\end{itemize}

We utilize the evaluation criteria outlined in \sec{\ref{sec:benchmark:eval}} for assessment.
DA tasks serve as a valuable lens to scrutinize the generalization capabilities of TSG methods across domains. 
%%%
It is pertinent to mention that for the generalization test, efficient processing of data from the target domain is essential, making the training efficiency of a TSG method a pivotal consideration. 
Thus, the results from DA can enhance other time series tasks by providing synthetic data for target domains with limited sample availability. 

%% file: tables/datasets.tex
\begin{table}[t]
\centering
\captionsetup{skip=0.75em}
\small
\caption{The statistics of the ten datasets.}
\label{tab:dataset}%
\begin{tabular}{lllll} \toprule
  \textbf{Datasets} & $R$ & $l$ & $N$ & \textbf{Domain} \\ 
  \midrule
  DLG \cite{misc_dodgers_loop_sensor_157} & 246 & 14 & 20 & Traffic \\
  Stock \cite{timegan} & 3,294 & 24 & 6 & Financial  \\
  Stock Long \cite{timegan} & 3,204 & 125 & 6 & Financial  \\
  Exchange \cite{lai2018modeling} & 6,715 & 125 & 8 & Financial \\
  Energy \cite{misc_appliances_energy_prediction_374} & 17,739 & 24 & 28 & Appliances \\
  Energy Long \cite{misc_appliances_energy_prediction_374} & 17,649 & 125 & 28 & Appliances \\
  EEG \cite{eeg} & 13,366 & 128 & 14 & Medical \\
  HAPT \cite{misc_human_activity_recognition_using_smartphones_240} & 1,514 & 128 & 6 & Medical \\
  Air \cite{zheng2015forecasting} & 7,731 & 168 & 6 & Sensor \\
  Boiler \cite{sasa} & 80,935 & 192 & 11 & Industrial \\
  \bottomrule
\end{tabular}
\end{table}

%% file: 07_setup.tex
\section{Experimental Setup}
\label{sec:setup}

% not to be exhaustive / Initial evaluation
The primary goal of \textsf{TSGBench} is to establish a rigorous, standardized framework for evaluating various TSG methods.
As such, we will not conduct an exhaustive evaluation, given time and resource constraints \cite{timevae, gtgan}. 
Instead, we concentrate on assessing recent, prevalent methods that come with publicly available code and have exhibited state-of-the-art performance on selected datasets.

\paragraph{Algorithms}
Our initial experiments evaluate the performance of ten pivotal TSG algorithms mentioned in \sec{\ref{sec:taxonomy:methods}}. These algorithms, carefully selected from academic literature, include: 
(1) pure GAN-based methods: RGAN \cite{rcgan}, TimeGAN \cite{timegan}, RTSGAN \cite{rtsgan}, COSCI-GAN \cite{cosci-gan}, and AEC-GAN \cite{aec-gan};
(2) pure VAE-based method: TimeVAE \cite{timevae} and TimeVQVAE \cite{timevqvae};
(3) mixed-type methods: GT-GAN \cite{gtgan}, LS4 \cite{ls4}, and Fourier Flow \cite{fourier}.

\paragraph{Datasets}
Our experiments draw on ten real-world datasets across diverse domains, as detailed in \sec{\ref{sec:benchmark:datasets}}. 
Moreover, we extend HAPT, Air, and Boiler for generalization tests, as described in \sec{\ref{sec:benchmark:da}}.

\paragraph{Evaluation Measures}
We employ twelve evaluation measures outlined in \sec{\ref{sec:benchmark:eval}} to make a thorough and systematic assessment of the TSG algorithms.
The lower value means the better performance.
Specifically, for DS and PS, we adopt two layers of LSTM; for C-FID, we adopt ts2vec \cite{ts2vec} as the backbone. For all evaluation measures, we repeat them five times and report their average results.

\paragraph{Experiments Environments}
All experiments are conducted on a machine with Intel\textsuperscript{\textregistered} Xeon\textsuperscript{\textregistered} Gold 6342 CPU @ 2.80GHz, 64 GB memory, and NVIDIA GeForce RTX 3090.

% methods
% shape of (# time points, # series) --> shape of (# samples, seq_len, # series)
% shape of (L, N) --> shape of (R, l, N)

\paragraph{Parameter Settings}
For RGAN, the number of hidden units for GANs is set to $4n$.
%%%
For TimeGAN, we use the suggested settings \cite{timegan} and adopt three-layer GRUs for the network architectures.
%%%
For RTSGAN, we adhere to its complete time series generation \cite{rtsgan} and set $\beta_1 = 0.9$ and $\beta_2 = 0.999$.
%%%
For COSCI-GAN, we set $\gamma=5$, employ MLP-based networks for the central discriminator, and follow other hyper-parameters from \cite{cosci-gan}.
%%%
For AEC-GAN, we set the context length
$l_c=4$ if $l=16$, 
$l_c=85$ if $l=24$, 
$l_c=25$ if $l=125$, 
$l_c=28$ if $l=128$, 
$l_c=56$ if $l=168$, and 
$l_c=64$ if $l=192$, and set the generation length $l_q = l - l_c$.
%%%
For TimeVAE, we set the latent dimension to 8 and the hidden layer sizes to 50, 100, and 200.
%%%
For TimeVQVAE, we adopt settings from \cite{timevqvae}, with $n\_fft=8$ and varying $max\_epochs \in \{2000, 10000\}$ for two training stages.
%%%
For GT-GAN, we use regular time series from \cite{gtgan} and set $P_{MLE} = 2$.
The absolute and relative tolerances for the generator are set to 0.001 and 0.01 for Energy and Energy Long, respectively, and 0.01 and 0.001 for other datasets. 
%%%
For LS4, we set the latent space dimension to 5 and configured the batch size to 512 for Air and Boiler and 1024 for the rest, optimizing GPU utilization.
%%%
For Fourier Flow, which is primarily designed for individual series, we follow its guidelines \cite{fourier} and adapt it for time series with $N>1$ by using DFT \cite{oppenheim1999discrete} to each dimension.
We configure the hidden size to 50 and set the number of flows to 3 for Stock and Stock Long and 5 for others.

%% file: 08_experiments.tex
\section{Results Analysis}
\label{sec:results}

\begin{figure*}[t]%
\centering%
\captionsetup{skip=0.75em,belowskip=0.0em}%
\includegraphics[width=0.99\textwidth]{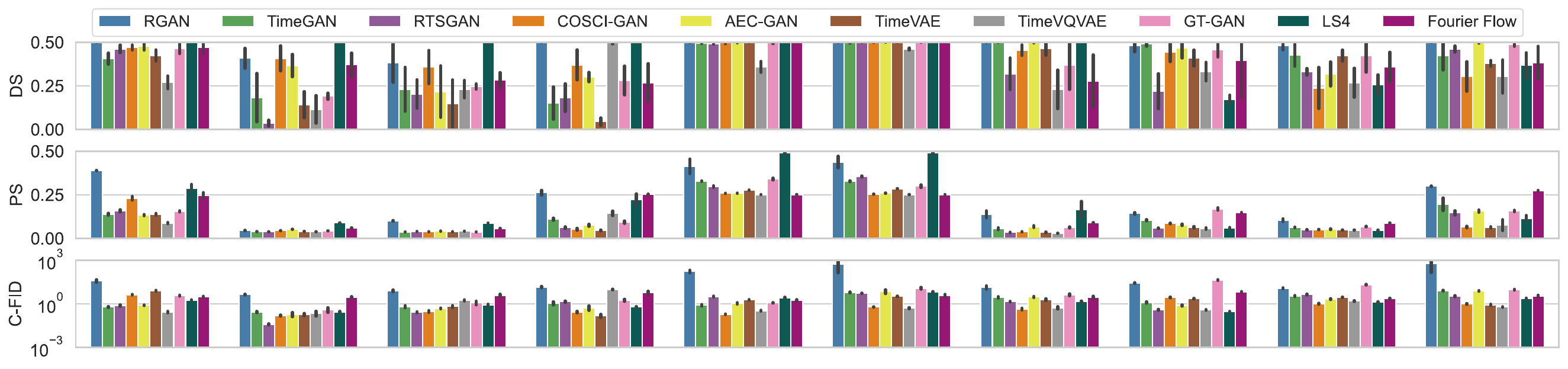}\\%
\includegraphics[width=0.99\textwidth]{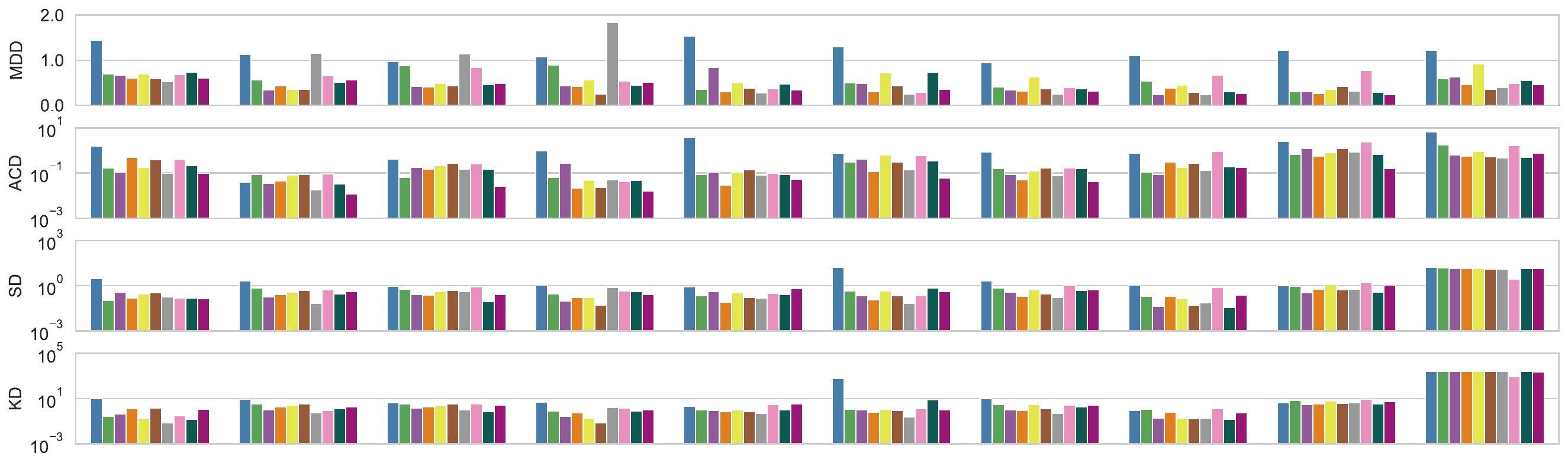}\\%
\includegraphics[width=0.99\textwidth]{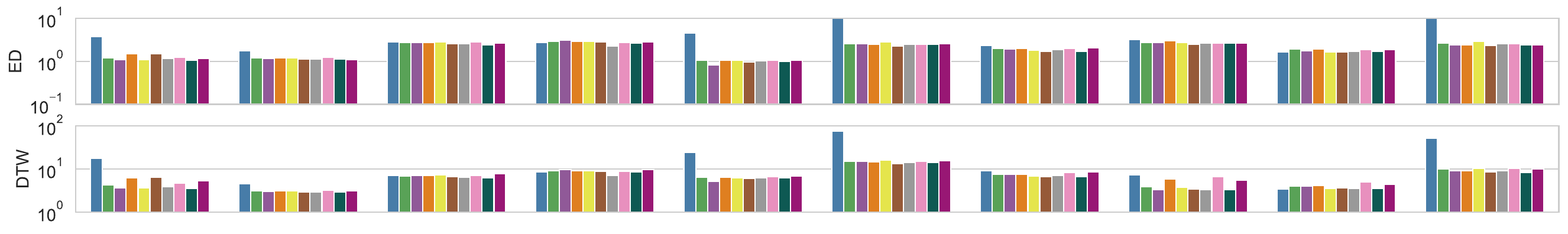}\\%
\includegraphics[width=0.99\textwidth]{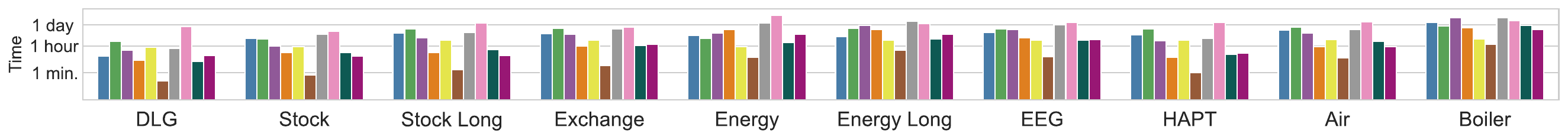}\\%
\caption{TSG benchmarking.}%
\label{fig:benchmarking}%
\end{figure*}%

\subsection{TSG Benchmarking}
\label{sec:results:benchmark}

\begin{figure*}[t]%
\centering%
\captionsetup{skip=0.55em,belowskip=0em}%
\includegraphics[width=0.98\textwidth]{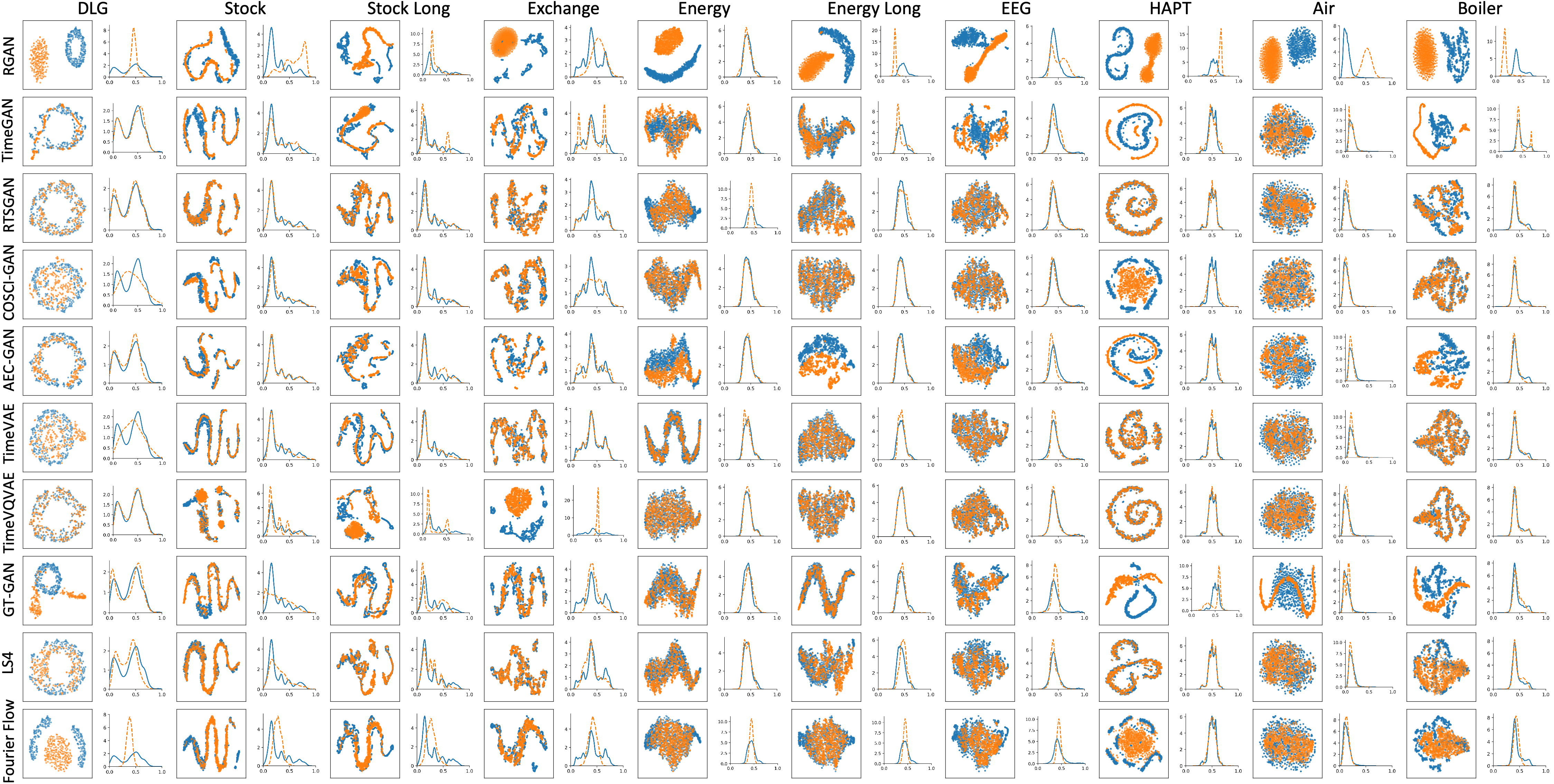}%
\caption{Visualization for TSG benchmarking by t-SNE and Distribution Plot (blue as $\bm{T}^{tr}$, orange as $\bm{T}^{gen}$).}%
\label{fig:vis_all}%
\end{figure*}

% Reports the quantitative and qualitative results
% Case study on opposite trend
We first present the results of ten methods applied to ten real-world datasets in Figures \ref{fig:benchmarking} and \ref{fig:vis_all}. 

\paragraph{Model-based Evaluations}
As evident from the first three rows of Figure \ref{fig:benchmarking}, TimeVQVAE, TimeVAE, RTSGAN, and COSCI-GAN consistently excel across the three model-based measures. 
In contrast, RGAN and TimeGAN underperform. 
The superior performance of VAE-based methods can be attributed to their adeptness at capturing temporal dependencies, pivotal for forecasting and representation learning tasks. 
Some methods like LS4, despite having high DS and PS, secure a relatively low C-FID. This indicates their satisfactory performance in the representation learning tasks.
%%%
Some outliers appear in Energy and Energy Long, where almost all methods achieve DS around 0.5. This suggests while the generated time series might suffice for predictive tasks, they are easily distinguishable in classification tasks.
%%%
Additionally, the large standard deviation in DS is worth noting, which will be further analyzed in \sec{\ref{sec:results:robust}}. 

%%% relations with dataset statistics 
Upon integrating observations from Table \ref{tab:dataset}, we find that as $l$ increases, PS will slightly drop, implying that shorter sequences present a greater challenge in generating time series. 
%%%
Furthermore, a relatively strong correlation appears between $N$ and these measures, suggesting that the generation of high-dimensional time series presents a steeper challenge, leading to higher DS and PS.

\paragraph{Feature-based Evaluations}
In Figure \ref{fig:benchmarking}, when examining feature-based measures from rows 4 to 7, Fourier Flow delivers the best performance in ACD, while COSCI-GAN dominates in MDD and SD. 
This is likely due to their effectiveness in capturing the statistical properties of time series.
%%%
Interestingly, within each dataset, the performance ranking across all four measures appears to be consistent. 
This consistency implies a robust correlation among these feature-based measures, indicating that the overall performance of a method in capturing key features is likely consistent as well.

%%% relations with dataset statistics
Moreover, feature-based measures, which primarily assess the statistical similarities between original and generated data, tend to improve when $N > 10$. This observation suggests that generated time series may better match the statistical properties of the original one when dealing with high-dimensional datasets, potentially due to the rich features available for modeling. 

\paragraph{Distance-based Evaluations}
The patterns observed in the distance-based measures, shown in the 8th and 9th rows of Figure \ref{fig:benchmarking}, differ from those in earlier measures. This divergence arises because distance-based measures direct quantification of dissimilarity between the generated and original time series. 
%%%
Regarding ED and DTW, VAE-based methods stand out. 
They effectively capture the overall trend of the original time series, preserving both value proximity (ED) and trend similarity (DTW).

%%% relations with dataset statistics
An intriguing finding is that the distances between generated and authentic time series amplify as $l$ increases. 
This is likely because longer sequences introduce more complex temporal dependencies, making them harder to model accurately, leading to larger divergence in values and alignment. 
%%%
Conversely, as $N$ grows, these distances decrease. The reason could be that a larger set of series provides a broader range of temporal patterns and inter-series correlations, benefiting the methods' evaluation.

\paragraph{Training Efficiency}
In the final row of Figure \ref{fig:benchmarking}, we explore the training efficiency. 
For ease of interpretation, we categorize training times into four distinct segments: $<1$ minute, $<1$ hour, $<1$ day, and $\geq 1$ day.
%%% fastest: VAE-based methods TimeVAE and LS4 
Our findings highlight the exceptional efficiency of TimeVAE and LS4, likely due to their VAE-based structure and efficient training strategies that minimize computational demands and iterations needed for convergence. 
%%%
On the other hand, TimeVQVAE, while effective in modeling, presents a more time-intensive training process. The complexity of TimeVQVAE arises primarily from its STFT process, tokenization, and iterative decoding. 
Also, our fairness-driven decision to maintain consistent hyper-parameters across all datasets potentially escalates training times for those with larger $R$ values. 
%%% slowest: GAN-based methods
In addition, GAN-based methods generally manifest longer training times. For example, GT-GAN takes training time of more than 1 day on all datasets except Stock, DLG, and Exchange. 
The inherent intricacy of GANs, requiring concurrent training of generator and discriminator networks to reach equilibrium, leads to a longer convergence period.

\paragraph{Visualization}
At last, we look into the t-SNE visualization and Distribution Plot of the generated time series in Figure \ref{fig:vis_all}. 
VAE-based methods, COSCI-GAN, and RTSGAN excel at generating time series that closely mirror the features and patterns of the original ones.
%%% analysis of outperforming methods
Examining the subtleties among these leading methods reveals variations in their generation capabilities for different distributions. 
For instance, the DLG dataset, characterized by its bimodal distribution, challenges COSCI-GAN, which struggles to capture both modes accurately. 
Conversely, TimeVAE and LS4 perform well with the Exchange dataset with the multifaceted peak structure, indicating their innate ability to grasp multifarious distributional patterns.

%%% from one distribution to another
Some methods, such as RGAN and GT-GAN, which might fare well on a single data distribution, cannot deal with the dramatic shift in the overall distributions (e.g., from Stock to HAPT). This suggests limitations in their ability to adapt to significantly different data distributions.
%%% similar distribution but cannot match t-sne
We also find that some methods, e.g., RGAN, while they successfully mimic the distribution of the original time series in their generated output (e.g., Energy), stumble when it comes to the precise matching by t-SNE. 
This observation aligns with the quantitative results in Figure \ref{fig:benchmarking}, underscoring the challenges in replicating exact values of the original time series rather than overall distribution characteristics.
%%% partial fit but with additional information
Moreover, some methods, like TimeGAN, can partially fit the original time series but usually contain extraneous information. 
They may struggle to handle the inherent noise present in real-world time series.

\begin{figure*}[t]%
\centering%
\captionsetup{skip=0.25em,belowskip=0em}%
\includegraphics[width=0.45\textwidth]{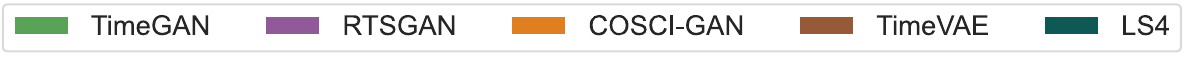}\\%
\subfigure[HAPT (Source: User 14).]{%
  \label{fig:da_hapt}%
  \includegraphics[width=0.495\textwidth]{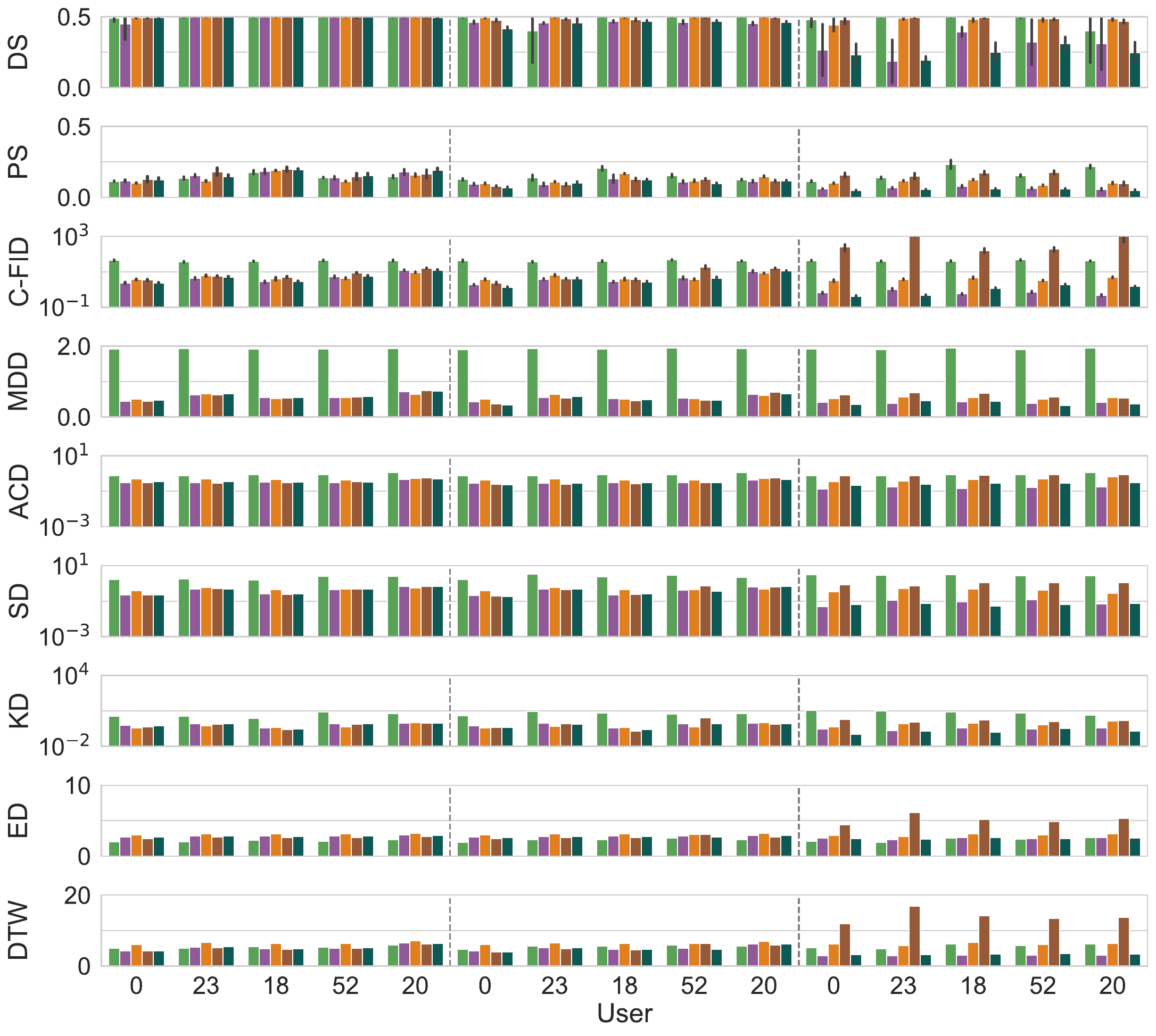}}%
\subfigure[Air (Source: Tianjin, TJ).]{%
  \label{fig:da_air}%
  \includegraphics[width=0.295\textwidth]{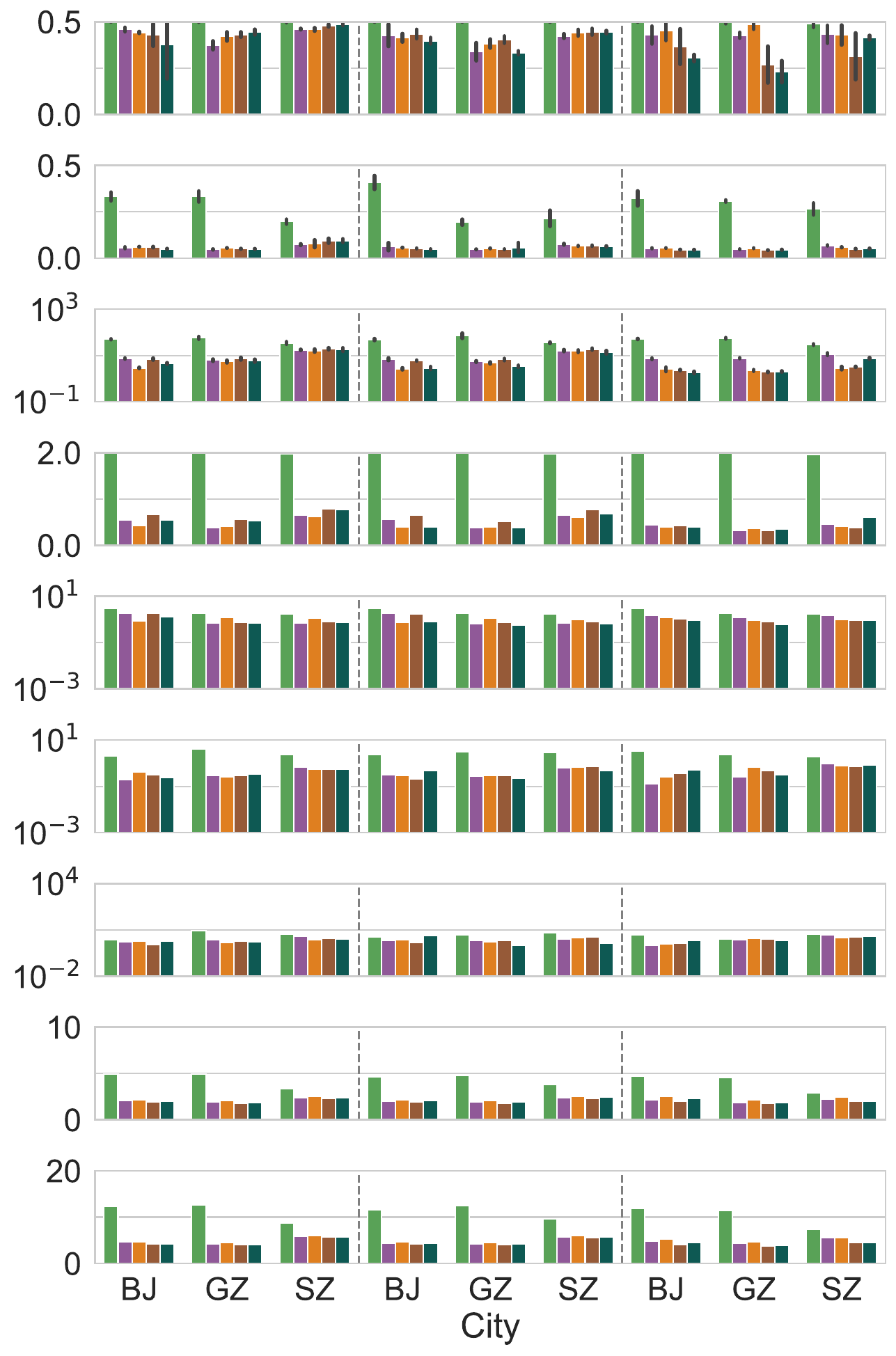}}%
\subfigure[Boiler (Source: Boiler 1).]{%
  \label{fig:da_boiler}%
  \includegraphics[width=0.195\textwidth]{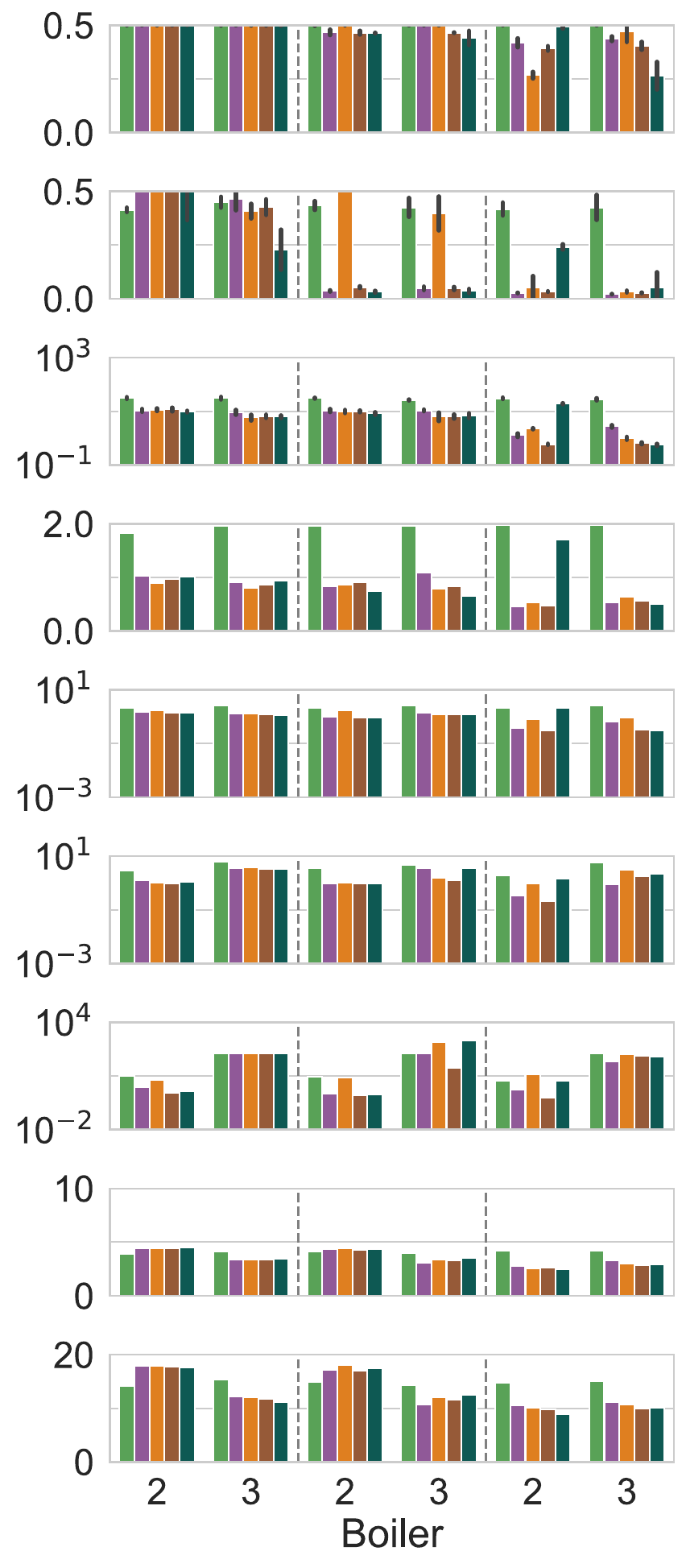}}%
\caption{Generalization test: single DA, cross DA, and reference DA (left to right in each subfigure).}%
\label{fig:da}%
\end{figure*}%

\subsection{Generalization Test}
\label{sec:results:da}

In conjunction with benchmarking results, we delve deeper into a select group of methods that showcased eminent performance, with a pronounced emphasis on efficiency, especially when contemplating generalization in the nascent stages of the target domain.
In particular, we employ TimeGAN as the baseline and focus on four efficient methods with leading performance, i.e., TimeVAE, COSCI-GAN, RTSGAN, and LS4. 
Quantitative results for the datasets HAPT, Air, and Boiler are displayed in Figure \ref{fig:da}, where each measure is partitioned into three parts by gray dashed lines, signifying the single DA, cross DA, and reference DA, respectively.

We first focus on the HAPT dataset, which consists of the most number of target domains.
% 1. some methods fail on DA
As depicted in Figure \ref{fig:da_hapt}, certain methods like TimeGAN show a little discrepancy in their performance across single DA, cross DA, and reference DA. This suggests that these methods may struggle with generalization in many cases. For instance, TimeGAN may struggle to adapt effectively due to reasons inherent to its model design. 
Another noteworthy observation is the considerable standard deviation seen in DS and PS under DA tasks, barring those that exhibit the worst score (0.5). This heightened variance undermines their reliability, which will be further analyzed in \sec{\ref{sec:results:robust}}.
% 2. v shape (two obs + two analysis)
When examining methods with strong generalization capabilities, the performance of some algorithms (e.g., TimeVAE and COSCI-GAN) in cross DA outperforms that in reference DA.
This indicates that they do not treat the smaller proportion of target domain time series $\bm{T}_t^{his}$ as noise, even when faced with multiple distributions as inputs.
Conversely, methods like RTSGAN and LS4 excel in single DA, as they quickly converge when given a limited set of time series. 
% 3. 20 vs. others
In addition, for the target domain User 20, TimeVAE's performance in reference DA matches the level achieved in cross DA. 
This may be due to User 20's data distribution being relatively simpler and less noisy, facilitating quicker convergence for TimeVAE during training

% effect of #samples and #dim
As $R$ and $N$ grow (from HAPT to Air and Boiler), we observe a consistent pattern for the single DA task in Figures \ref{fig:da_air} and \ref{fig:da_boiler}. In the case of the reference DA task, a larger $R$ facilitates model convergence, yielding more competitive outcomes compared to the cross DA task.
% effect of data distributions
Another interesting observation is the inconsistent performance of SD, KD, and DTW measures in evaluating DA for the Boiler dataset compared to the HAPT and Air datasets. This discrepancy is due to the periodic trends present in HAPT and Air. 
Since SD and KD gauge data distribution and DTW assesses alignment, they are less effective for datasets lacking periodic trends.

\subsection{Robustness Test for Evaluation Measures}
\label{sec:results:robust}

% Motivation
In the course of TSG benchmarking (\sec{\ref{sec:results:benchmark}}) and generalization testing (\sec{\ref{sec:results:da}}), we observed significant variability and inconsistency in DS and PS. 
Thus, we undertook additional analyses to evaluate the sensitivity and robustness of various evaluation measures.

% Settings: by synthetic input and output, to test the upper and lower bound of the measures.
We randomly generated 10,000 synthetic time series with $N=5$ adhering to the sine function \cite{timegan}, expressed as $x_{i,j} = \sin (2 \pi \eta j+\theta)$, where $\eta \sim \mathcal{U}[0,1]$, $\theta \sim \mathcal{U}[-\pi, \pi]$, $i \in [1,5]$, and $j \in [1,l]$. 
We then assessed these series using two sequence lengths, $l=24$ and $l=125$, employing the evaluation measures outlined in \sec{\ref{sec:benchmark:eval}}. 
The predictive score incorporated two configurations: next step forecasting \cite{timegan} (denoted as PS) and entire sequence forecasting (denoted as PS (entire)) \cite{gtgan}.
Table \ref{tab:two-scores} considered two scenarios for input time series: (1) identical original and generated data, where ideal evaluation measures should be 0; and (2) time series sampled from the same sine function but with different seeds.

% Results
%%% 1. identical vs. random sampling
In Table \ref{tab:two-scores}, it is evident that all feature-based, distance-based measures and C-FID demonstrate the robustness and accurate reflection of the changes in input time series.
Contrarily, DS and PS exhibit some inconsistencies. 
For instance, the scores for DS and PS from random sampling at $l=125$ manifest lower values than their identical counterparts, a result that tends to be counter-intuitive.
Also, the standard deviation of DS is notably high relative to the mean value, suggesting considerable variation.
%%% 2. insensitive to l: 24 vs. 125
Moreover, DS appears insensitive to the increase in $l$, limiting its effectiveness in accurately evaluating the TSG methods' performance on longer sequences.
%%% 
The above observations confirm that DS and PS may be less robust than other measures, especially as input settings change. This aligns with the challenges we discussed in \sec{\ref{sec:intro:motivation}}.

\input{tables/robustness}

\subsection{Ranking Analysis}
\label{sec:results:ranking}

Selecting suitable TSG methods is vital for handling new time series datasets. Therefore, it is crucial to understand and analyze the consistency in performance across various methods.

\paragraph{Method Ranking}
We initially present the ranking of ten methods under two specific scenarios. 
First, we evaluate their performance across all datasets for each individual measure, as depicted on the left-hand side of Figure \ref{fig:rank_all}. 
Second, we examine their average ranking across all measures but constricted to each dataset included in \textsf{TSGBench}, as showcased on the right-hand side of Figure \ref{fig:rank_all}.

%%% best and worst --> consistent
Across all evaluation measures and datasets, no single method consistently dominates, but TimeVQVAE, TimeVAE, COSCI-GAN, RTSGAN, and LS4 often outperform others. 
Specifically, TimeVAE and LS4 excel in distance-based measures and demonstrate impressive training efficiency, while COSCI-GAN and RTSGAN lead in model-based measures. 
In contrast, RGAN generally ranks lower in performance. 
%%%
A dataset-centric analysis reveals a similar pattern, with the same group of TimeVQVAE, TimeVAE, COSCI-GAN, RTSGAN, and LS4 achieving high rankings across various datasets. This consistency across different evaluation aspects further validates the robustness and reliability of \textsf{TSGBench}.
% When we pivot our focus to dataset-centric performance, a similar narrative unfolds. The aforementioned quartet of TimeVQVAE, TimeVAE, COSCI-GAN, RTSGAN, and LS4 continues to secure favorable rankings across various datasets. The consistency in rankings observed across the two evaluative dimensions reaffirms the robustness and credibility of \textsf{TSGBench}.

\begin{figure}[t]%
\centering%
\captionsetup{skip=0.75em,belowskip=0em}%
\includegraphics[width=0.99\columnwidth]{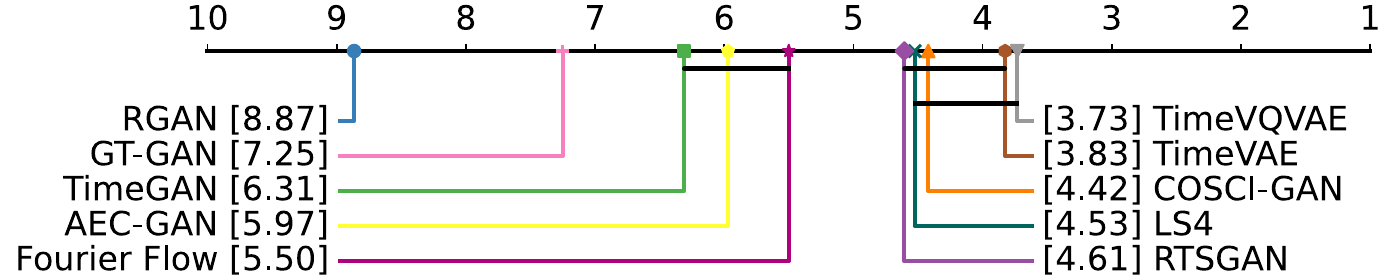}%
\caption{Critical difference diagram of TSG methods.}%
\label{fig:cd}%
\end{figure}

% Statistical tests of ranks
\paragraph{Statistical Validation}
To statistically confirm the method ranking, we employ the Friedman test \cite{friedman1937use} along with Conover's test \cite{conover1979multiple, Terpilowski2019} for ranking comparisons.
Figure \ref{fig:cd} presents the average rankings and the critical difference for each method.

% four groups
The ten TSG methods can be segregated into four tiers. TimeVQVAE, TimeVAE, COSCI-GAN, LS4, and RTSGAN lead, followed by Fourier Flow, AEC-GAN, and TimeGAN. GT-GAN forms the third cohort, while RGAN occupies the lowest tier. This aligns with the observations from \sec{\ref{sec:results:benchmark}}.
% crossbar
The supremacy of the first-tier methods becomes unambiguously clear as they stand statistically distinct from their counterparts spanning the other three tiers. 
%%%
Zooming into this premier group, despite the noticeable lead of TimeVQVAE and TimeVAE, they do not statistically tower over their peers within the group. 
%%%
This statistical overlap is not confined to the elite group but is also exhibited in the second group. Such an overlap indicates that the performance deltas between methods within these groups are not glaringly vast. 
Conversely, the third and fourth groups' distinct position underscores a pronounced performance chasm, setting it distinctly apart from others.

\subsection{Recommendations}
\label{sec:results:recommendations}

Finally, we provide guidelines to assist users in effectively using \textsf{TSGBench}. 
This benchmark is designed to be a beacon for navigating the intricate terrain of TSG.

\paragraph{Selection of TSG Methods} 
\textsf{TSGBench} serves as a toolkit to effectively discern the most appropriate TSG methods for different datasets. 
When confronted with a new dataset, an insightful strategy would be to juxtapose the statistical properties and distributions of the new time series datasets against those cataloged in \textsf{TSGBench}. 
This strategy offers a navigational compass, pointing users towards relevant generation techniques.
\begin{enumerate}[nolistsep,leftmargin=25pt]
  \item As a foundational step, we advocate for users to commence with VAE-based methods (e.g., TimeVAE and LS4). Their consistent leading performance and superior computational efficiency make them go-to choices for initial exploration. 

  \item In applications emphasizing autocorrelation or forecasting, such as predictive maintenance or stock market analysis, the ACD measure becomes crucial. 
  Fourier Flow, which is recognized for maintaining temporal dependencies, is highly suitable for these scenarios. 
  On the other hand, for capturing complex multi-variate relationships in datasets, COSCI-GAN is the recommended choice. 

  \item Subsequent considerations focus on dataset size and domain specificity. 
  For small-sized datasets, RTSGAN and LS4, which excel in single DA, are strong choices. 
  For heterogeneous datasets, or when the goal is to generate time series for a new target domain, TimeVAE and COSCI-GAN stand out for their effectiveness in cross DA.
  
  \item Users can further fine-tune their method selection based on specific real-world application needs, which involves identifying the most relevant evaluation measures. 
  In this case, Figure \ref{fig:rank_all} serves as a valuable visual guide.
\end{enumerate}

\paragraph{Selection of Evaluation Measures}
When evaluating a new TSG method, a comprehensive assessment is essential. 
Leveraging the features of \textsf{TSGBench}, we offer the following guidelines to facilitate this process. 
This allows users to tailor their choice of evaluation measures to the specific application requirements.
\begin{enumerate}[nolistsep,leftmargin=25pt]
  \item For applications where generated series will be used in classification or forecasting, model-based measures are advisable. Nevertheless, considering the robustness issues with DS and PS, we recommend starting with C-FID. 
  Its prowess in gauging fidelity based on representations bestows it with the capability to augment subsequent tasks. 
  
  \item When the goal is to emphasize the statistical attributes of the dataset, feature-based measures emerge as the preferred option, offering precise insights into the statistical nuances.
  % If the intent is to foreground statistical characteristics, feature-based measures become the natural choice. 
  
  \item In projects focusing on time series clustering, distance-based metrics assume an elevated importance due to their ability to discern subtle distinctions and similarities within the data. 
\end{enumerate}

Thus, users can customize their choice of evaluation measures to align closely with their research goals, ensuring more robust results.
Note that not all measures tend to yield uniform outcomes consistently. Users could continually calibrate their evaluations and discern the trade-off between effectiveness and efficiency. This ensures that the chosen measures are not only reflective of the desired outcomes but also optimized for performance.

%% file: tables/robustness.tex
\begin{table*}[t]
\centering
\captionsetup{skip=0.75em}
\caption{Robustness test on ten evaluation measures.}
\label{tab:two-scores}%
\resizebox{\textwidth}{!}{%
  \begin{tabular}{cccccccccccc}
  \toprule
  \multirow{1.5}[4]{*}{\textbf{Input}} & \multirow{1.5}[4]{*}{\textbf{Shape} $(R,l,N)$} & \multicolumn{4}{c}{\textbf{Model-based}} & \multicolumn{4}{c}{\textbf{Feature-based}} & \multicolumn{2}{c}{\textbf{Distance-based}} \\
  \cmidrule(lr){3-6}\cmidrule(lr){7-10}\cmidrule(lr){11-12} 
  & & \textbf{DS} & \textbf{PS}  & \textbf{PS (entire)} & \textbf{C-FID} & \textbf{MDD} & \textbf{ACD} & \textbf{SD} & \textbf{KD} & \textbf{ED} & \textbf{DTW} \\
  \midrule
  \multirow{1.5}[2]{*}{Identical} 
  & (10,000,~24,~5)  & 0.006±0.003 & 0.094±0.000 & 0.072±0.005 & 0.000±0.000 & 0.001 & 0.000 & 0.000 & 0.000 & 0.000 & 0.000 \\
  & (10,000,~125,~5) & 0.010±0.007 & 0.251±0.003 & 0.169±0.001 & 0.000±0.000 & 0.000 & 0.000 & 0.000 & 0.000 & 0.000 & 0.000 \\
  \multirow{1.5}[2]{*}{Random Sampling} 
  & (10,000,~24,~5)  & 0.009±0.005 & 0.094±0.000 & 0.071±0.005 & 0.003±0.000 & 0.222 & 0.131$\times10^{-3}$ & 0.009 & 0.007 & 0.653 & 1.689 \\
  & (10,000,~125,~5) & 0.003±0.005 & 0.249±0.003 & 0.168±0.001 & 0.016±0.001 & 0.108 & 0.022 & 0.009 & 0.020 & 4.350 & 9.663 \\
  \bottomrule
  \end{tabular}%
}
\end{table*}

% STA ACF: 0.013, 0.058

%% file: 09_conclusions.tex
\section{Conclusion and Future Work}
\label{sec:conclusions}
In this paper, we propose \textsf{TSGBench}, a groundbreaking benchmark specifically designed for TSG.
%%% benchmark
This benchmark is comprehensive, featuring datasets from diverse domains, a standardized data preprocessing pipeline, a holistic evaluation suite, and a novel generalization test grounded in DA.
%%% results
Extensive results validate its capability to offer a unified and equitable platform for assessing the efficacy and robustness of various TSG methods. Importantly, \textsf{TSGBench} also sheds light on the potential of their generalization capabilities.
%%% impacts
As a collaborative resource, it promises to catalyze further advancements within the time series community.

%%% Future Work
Looking forward, our aspirations for \textsf{TSGBench} are multifold. We intend to continually integrate emerging TSG methods, ensuring the benchmark remains at the vanguard of advancements. 
Also, the addition of new datasets is on the horizon, aiming to enhance the diversity and complexity of the challenges the benchmark addresses. 
Lastly, we are also contemplating introducing functionalities that facilitate automatic tuning, thereby streamlining the training process and making it even more friendly for users.

\begin{acks}
This research is supported by the National Research Foundation, Singapore under its Strategic Capability Research Centres Funding Initiative and the Ministry of Education, Singapore, under its MOE AcRF TIER 3 Grant (MOE-MOET32022-0001).
Any opinions, findings and conclusions or recommendations expressed in this material are those of the author(s) and do not reflect the views of National Research Foundation, Singapore and the Ministry of Education, Singapore.
\end{acks}

% %%% N-CRiPT
% This research is supported by the National Research Foundation, Singapore under its Strategic Capability Research Centres Funding Initiative. 
% Any opinions, findings and conclusions or recommendations expressed in this material are those of the author(s) and do not reflect the views of National Research Foundation, Singapore.

% %%% iGYRO
% This research/project is supported by the Ministry of Education, Singapore, under its MOE AcRF TIER 3 Grant (MOE-MOET32022-0001).
% Any opinions, findings and conclusions or recommendations expressed in this material are those of the author(s) and do not reflect the views of the Ministry of Education, Singapore.